\documentclass{article}

\usepackage[utf8]{inputenc}
\usepackage[T1]{fontenc}    
\usepackage{url}            
\usepackage{booktabs}      
\usepackage{amsfonts}       
\usepackage{nicefrac}       
\usepackage{microtype}      
\usepackage{xcolor}         
\usepackage[table]{xcolor}

\usepackage{kotex}

\usepackage{caption}
\usepackage{makecell}
\usepackage{multirow}
\usepackage{multicol}
\usepackage{colortbl}
\usepackage{subcaption}
\usepackage{lipsum,calc}
\newsavebox{\leftbox}
\newsavebox{\rightbox}
\usepackage{tablefootnote}
\usepackage{pifont}

\usepackage{wrapfig}
\usepackage{graphicx}
\usepackage{amsmath}
\usepackage{ulem}
\usepackage{hyperref}
\usepackage{pifont}

\usepackage[final]{corl_2026}

\title{\LARGE \bf
FlowVLA: Enabling World Dynamics Understanding through Pluggable 3D Flow Learning
}
\author{Junhyeong Go$^{1}$\qquad Mingyu Jung$^{1}$\qquad Woobin Im$^{2}$\qquad Jongbin Ryu$^{1}$
\\[0.4em]
{\small
$^{1}$Ajou University
\qquad
$^{2}$Samsung Electronics
}
\\[0.4em]
\href{https://mmai-laboratory.github.io/flowvla/}{\textcolor[HTML]{FF0080}{\textbf{https://mmai-laboratory.github.io/flowvla/}}}
}

\begin{document}
\maketitle

\newcommand{\rebfig}{
\begin{figure}[h]
  \centering
  \includegraphics[width=1\linewidth]{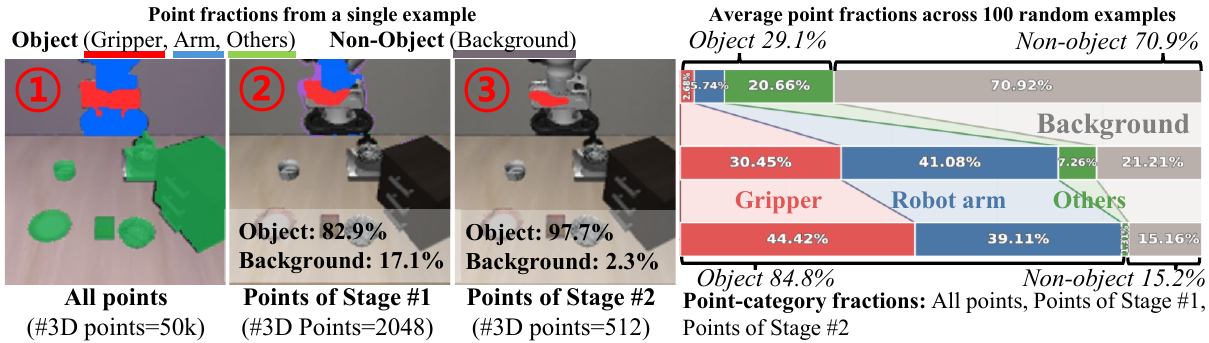}
  \caption{\textbf{3D point composition analysis.} Motion-based selection increases the proportion of object points, while TCP aggregation further reduces background points.}
   \label{figure:3dpoints}
\end{figure}
}

\newcommand{\figOverall}{
\begin{figure*}[t!]
    \centering
    \includegraphics[width=1.0\textwidth, height=0.5\textwidth]{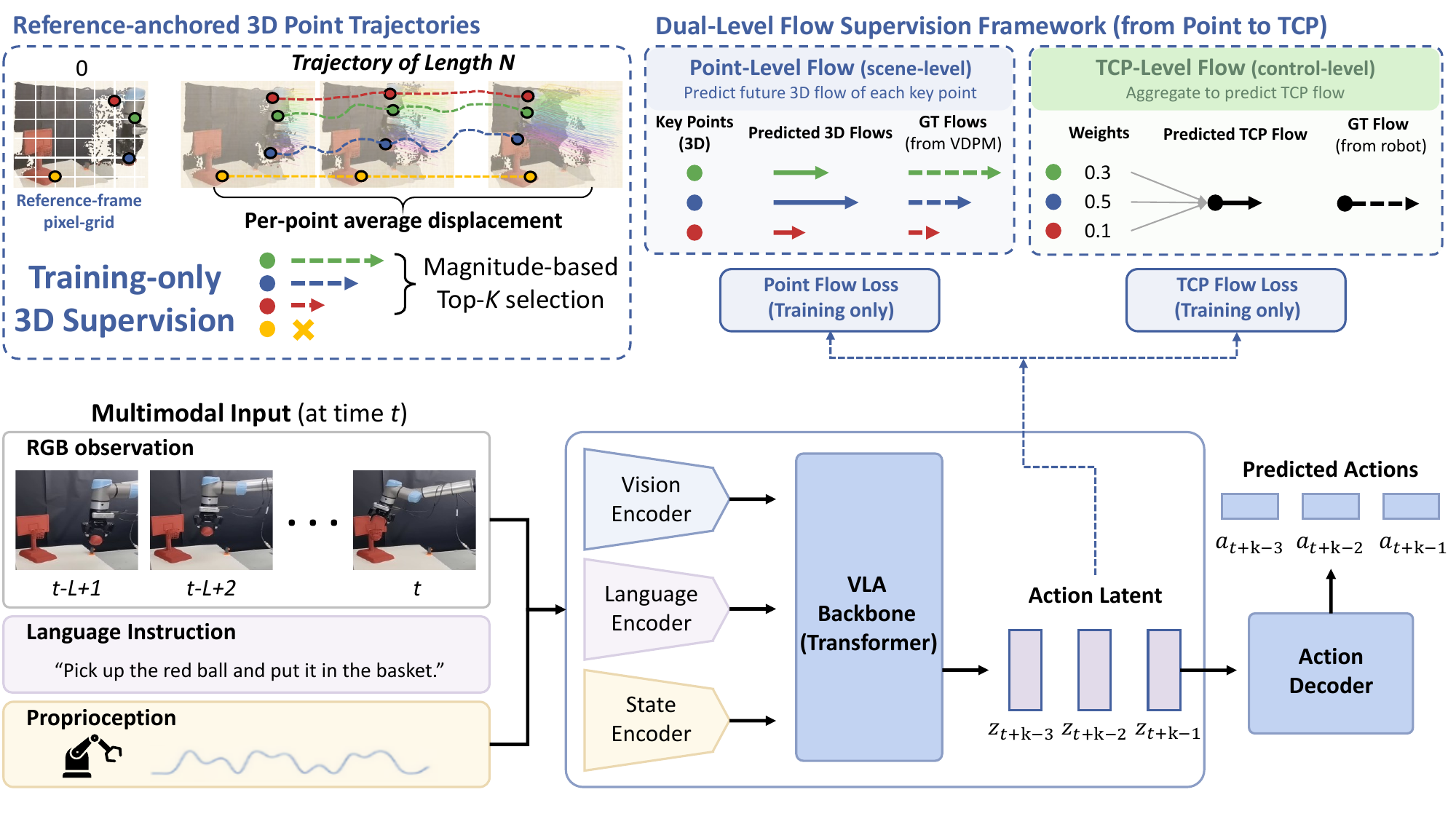}
    \caption{
   \textbf{Overview of Training-only 3D flow supervision.} Reference-anchored 3D points are tracked over time to construct 3D flow targets, from which top-$K$ motion-salient points are selected. Point-Level Flow predicts scene-level motion, while TCP-Level Flow aggregates it into end-effector motion, bridging scene-level dynamics and control-relevant motion.
    } 
    \label{figure:overall}
    
\end{figure*}
}

\newcommand{\figliberoplus}{
\begin{figure*}[h!]
    \centering
    \includegraphics[width=1.0\textwidth, height=0.57\textwidth]{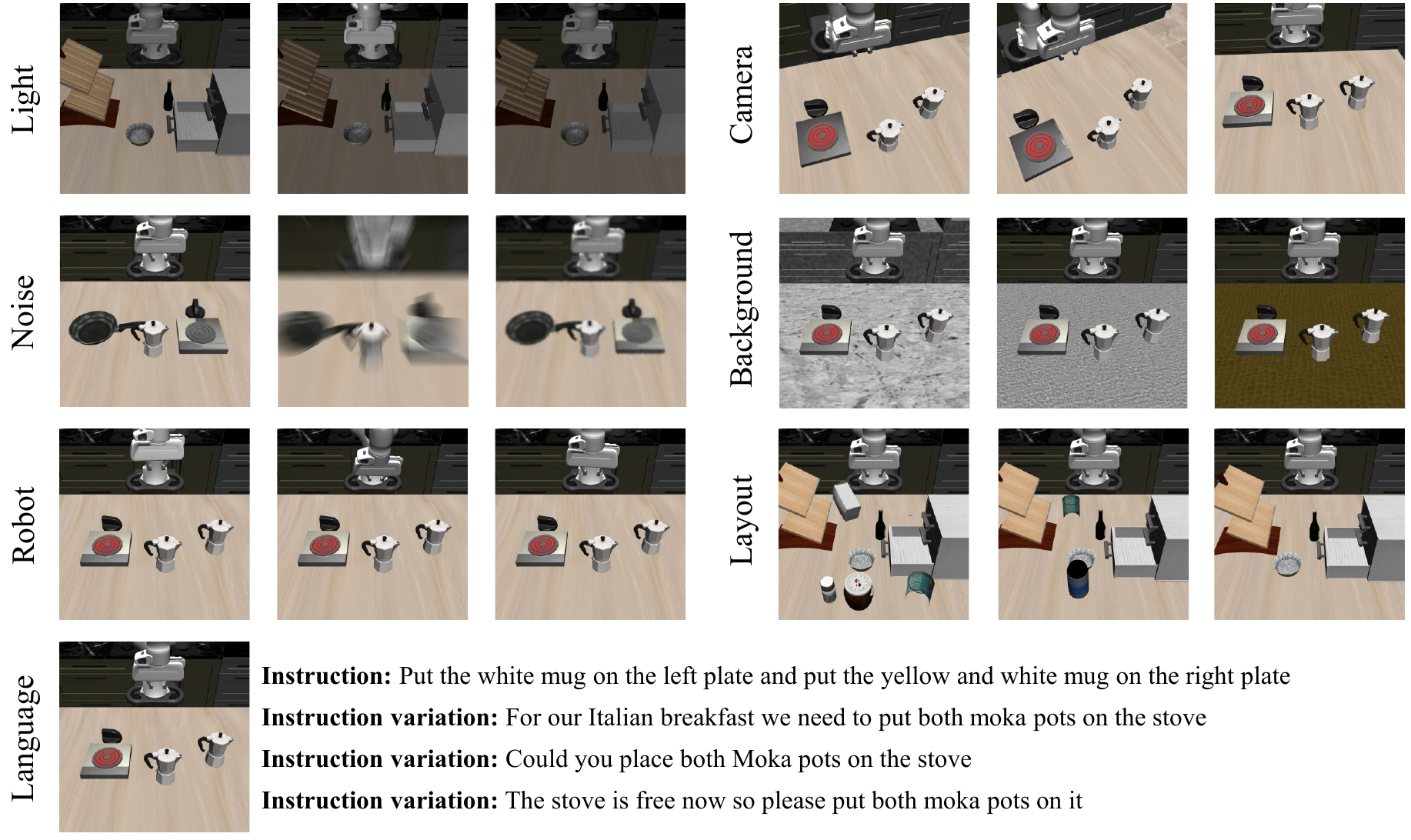}
\caption{\textbf{LIBERO-Plus Benchmark.} LIBERO-Plus evaluates robustness under distribution shifts across seven perturbation categories: lighting, camera viewpoint, sensor noise, background, language instruction, scene layout, and robot initialization.}
\label{figure:libero_plus_setting}
\vspace{-4mm}
\end{figure*}
}

\newcommand{\figvdpmvisualize}{
\begin{figure*}[h!]
    \centering
    \includegraphics[width=1.0\textwidth, height=0.37\textwidth]{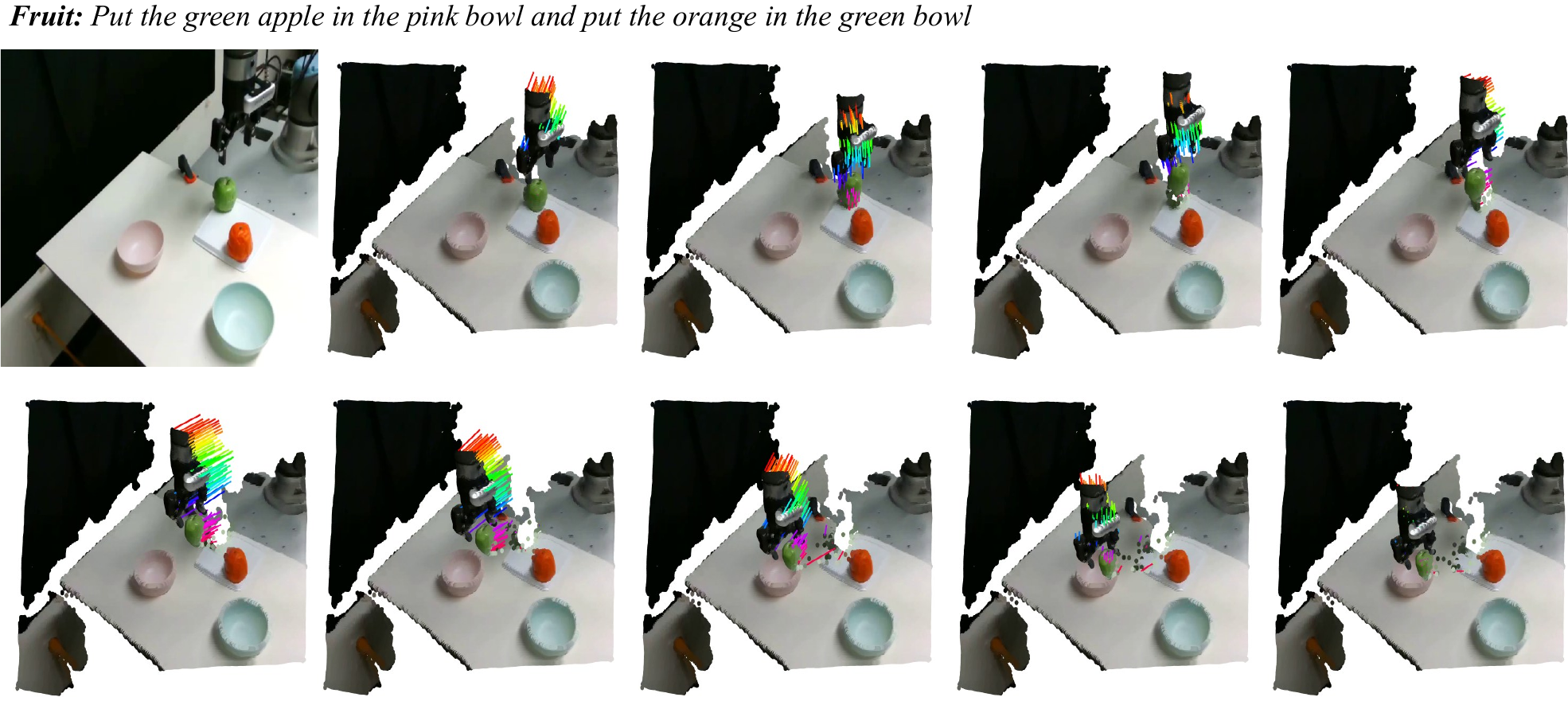}
    \caption{
   \textbf{VDPM 3D point trajectory visualization} on the real-world Fruit task.
    } 
    \label{figure:vdpm_visualize}
\end{figure*}
}

\newcommand{\figvdpmvisualizelibero}{
\begin{figure*}[h!]
    \centering
    \includegraphics[width=1.0\textwidth, height=0.37\textwidth]{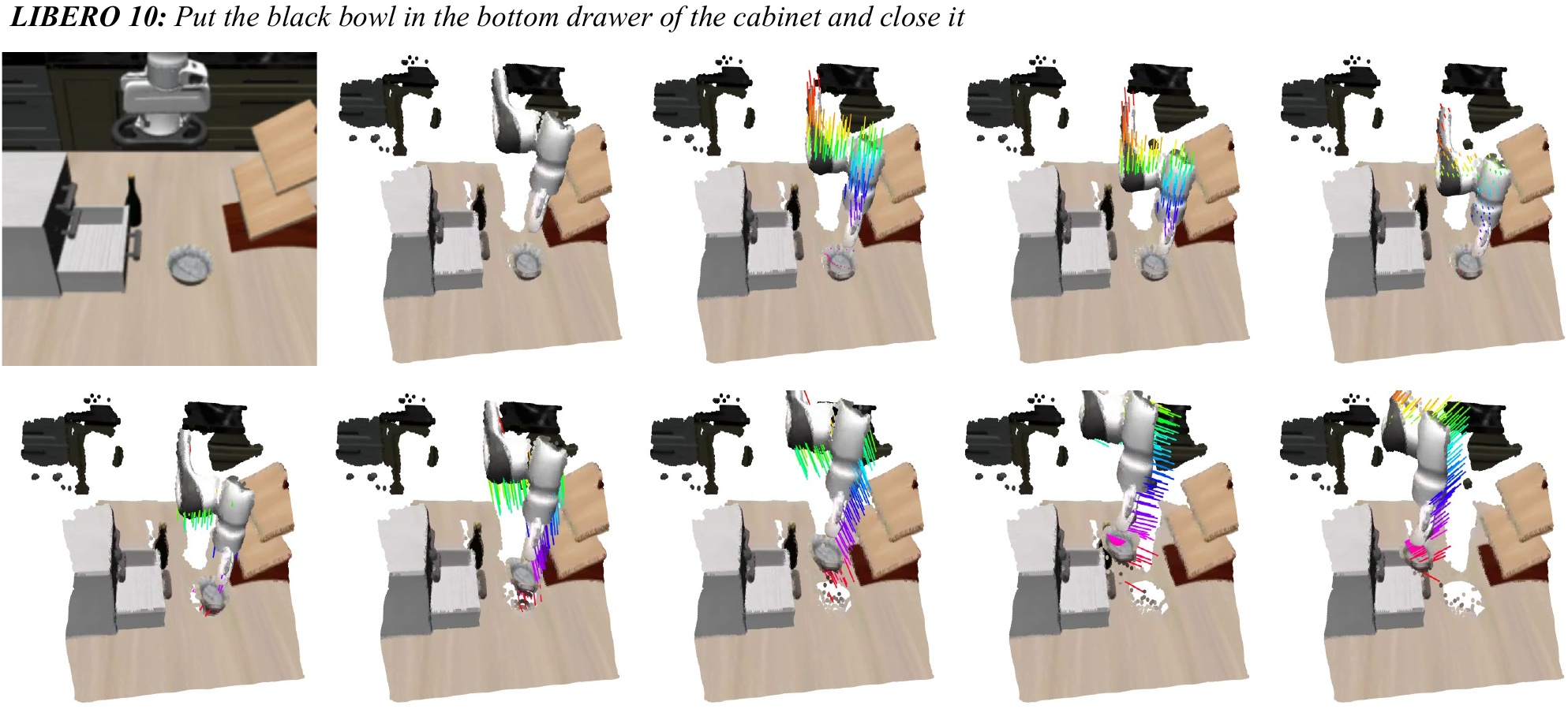}
    \caption{
   \textbf{VDPM 3D point trajectory visualization} on the LIBERO-10 task.
    } 
    \label{figure:vdpm_visualize_libero}
\end{figure*}
}

\newcommand{\figmagnitudefiltering}{
\begin{figure*}[h!]
    \centering
    \includegraphics[width=1.0\textwidth]{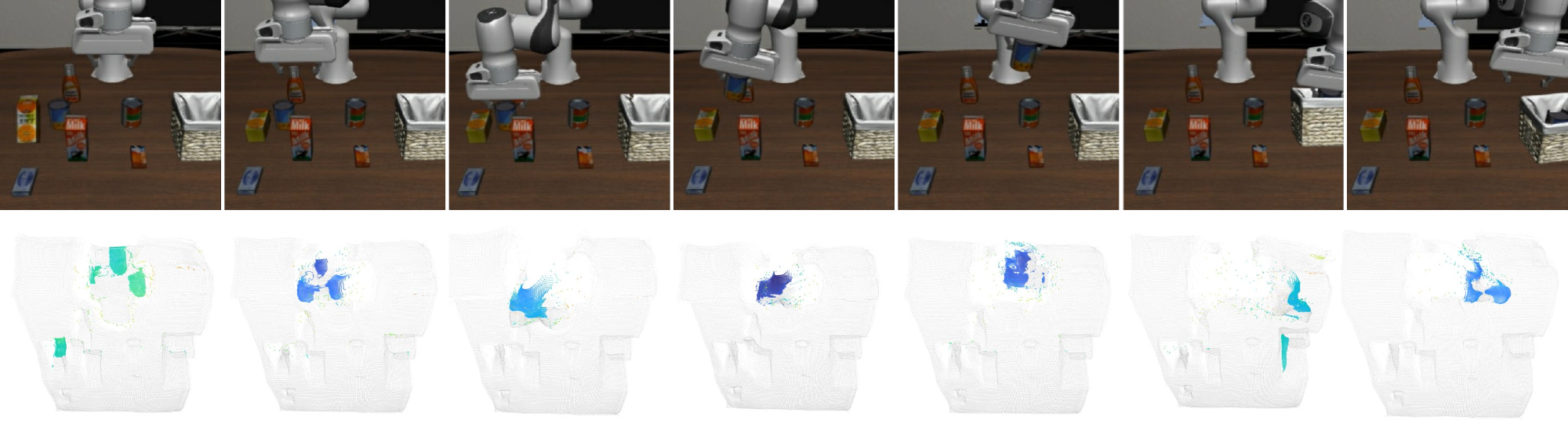}
    \caption{
   \textbf{Qualitative visualization} of magnitude-based point selection.
    } 
    \label{figure:magnitude_viz}
\end{figure*}
}

\newcommand{\figaggregatedweight}{
\begin{figure*}[h!]
    \centering
    \includegraphics[width=1.0\textwidth]{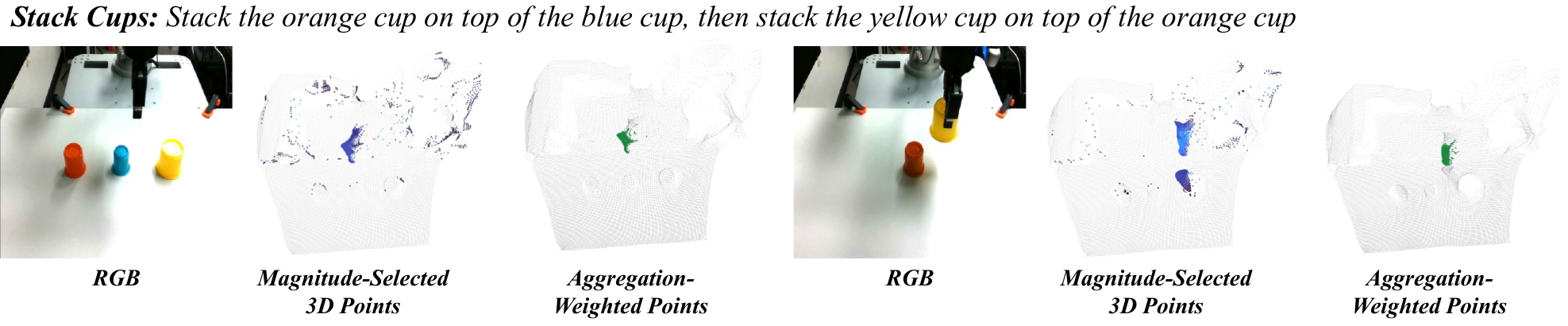}
    \caption{
   \textbf{Qualitative visualization} of TCP-level aggregation weights.
    } 
    \label{figure:tcpaggregate_viz}
\end{figure*}
}

\newcommand{\figtcptrajectory}{
\begin{figure*}[h!]
    \centering
    \includegraphics[width=1.0\textwidth]{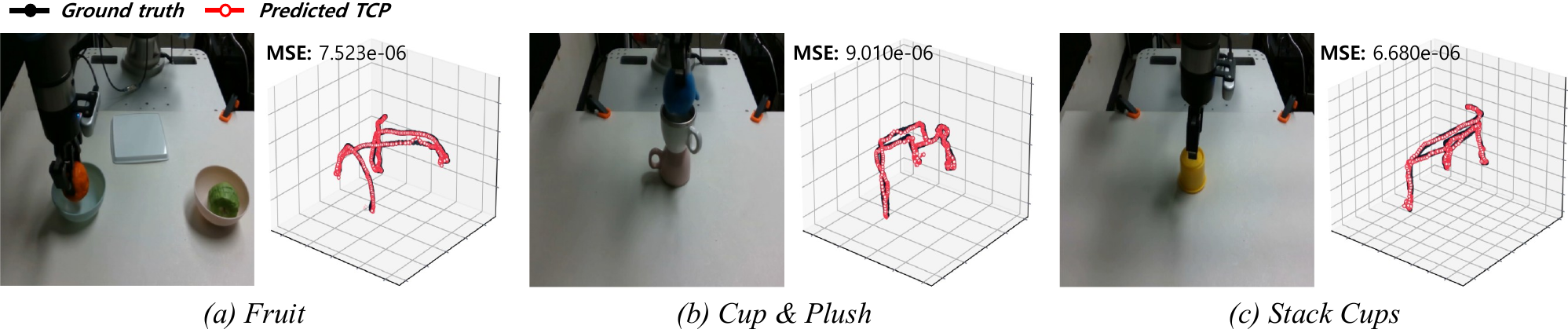}
    \caption{
   \textbf{Qualitative visualization} of predicted TCP trajectory.
    } 
    \label{figure:tcp_trajectory_viz}
    \vspace{-3mm}
\end{figure*}
}

\newcommand{\figRealcasesecond}{
\begin{figure*}[h!]
    \centering
    \includegraphics[width=1.0\textwidth, height=0.6\textwidth]{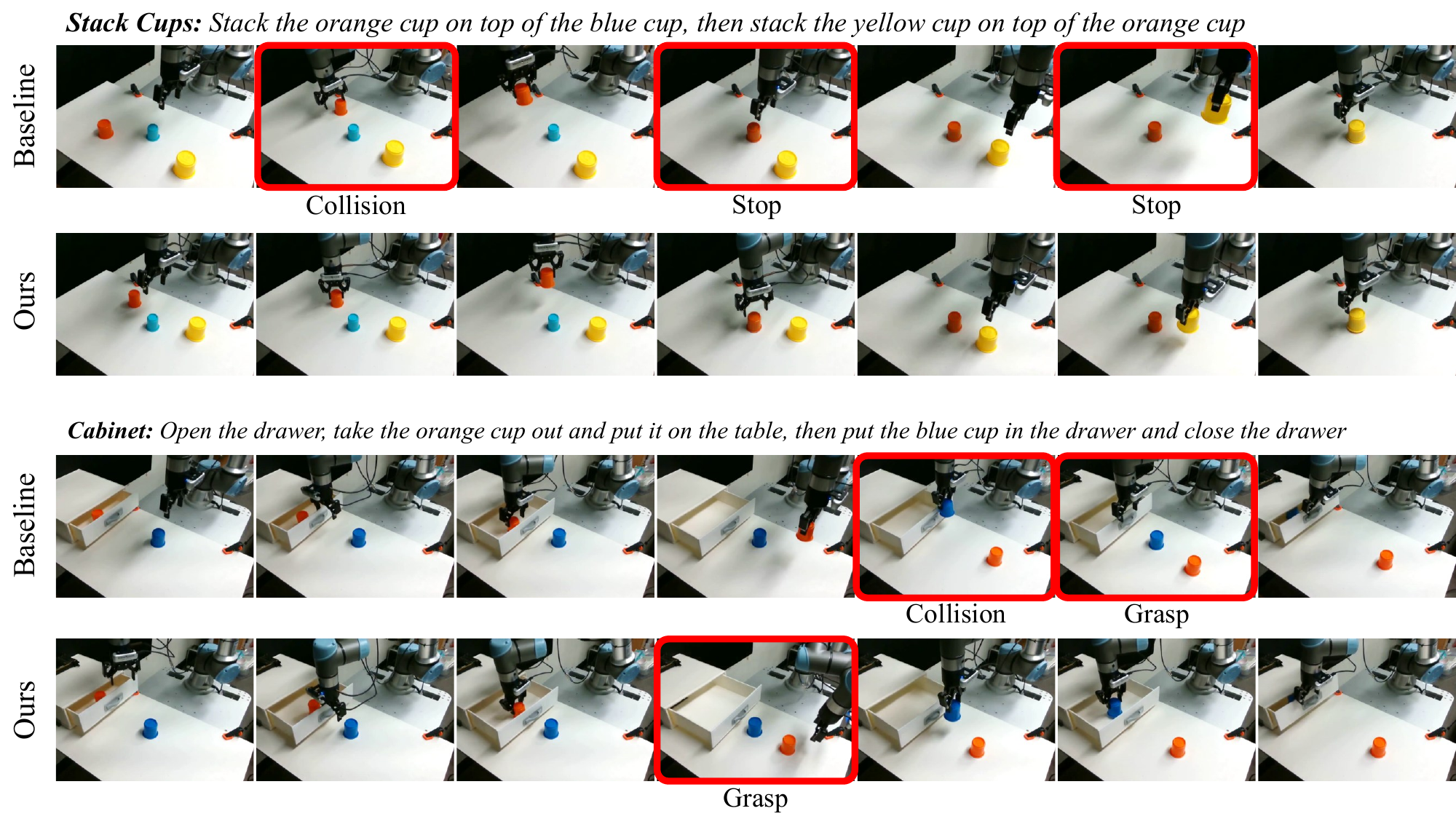}
    \caption{
   \textbf{Failure case visualization} on \textit{Stack Cups} and \textit{Cabinet}. 
    } 
    \label{figure:failurecase_basic_part2}
\end{figure*}
}

\newcommand{\figRealcasebasket}{
\begin{figure*}[h!]
    \centering
    \includegraphics[width=1.0\textwidth, height=0.6\textwidth]{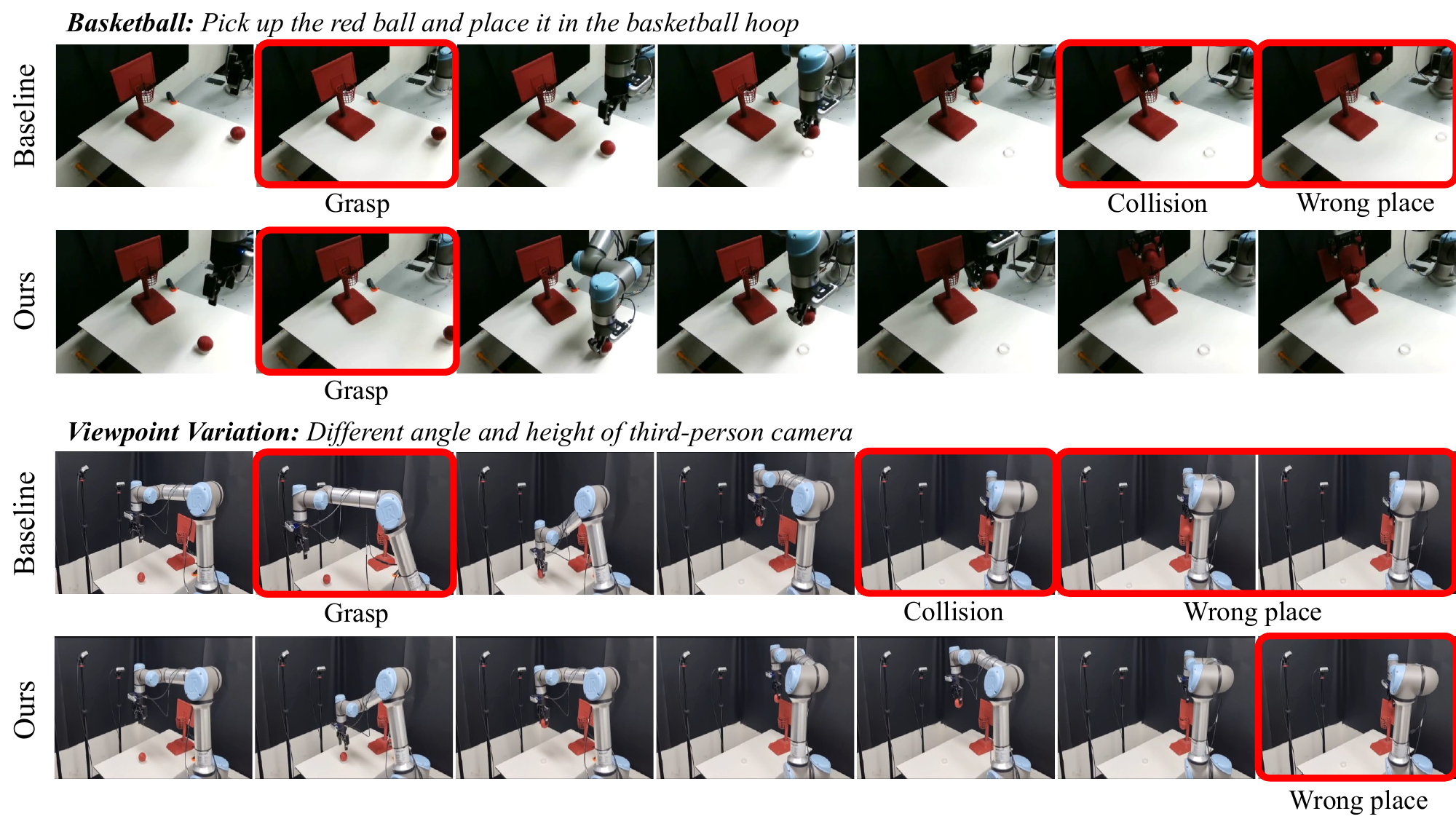}
    \caption{
   \textbf{Failure case visualization} on \textit{Basketball} with \textbf{viewpoint variation}. 
    } 
    \label{figure:failurecase_basketball}
\end{figure*}
}

\newcommand{\figRealcasering}{
\begin{figure*}[h!]
    \centering
    \includegraphics[width=1.0\textwidth, height=0.6\textwidth]{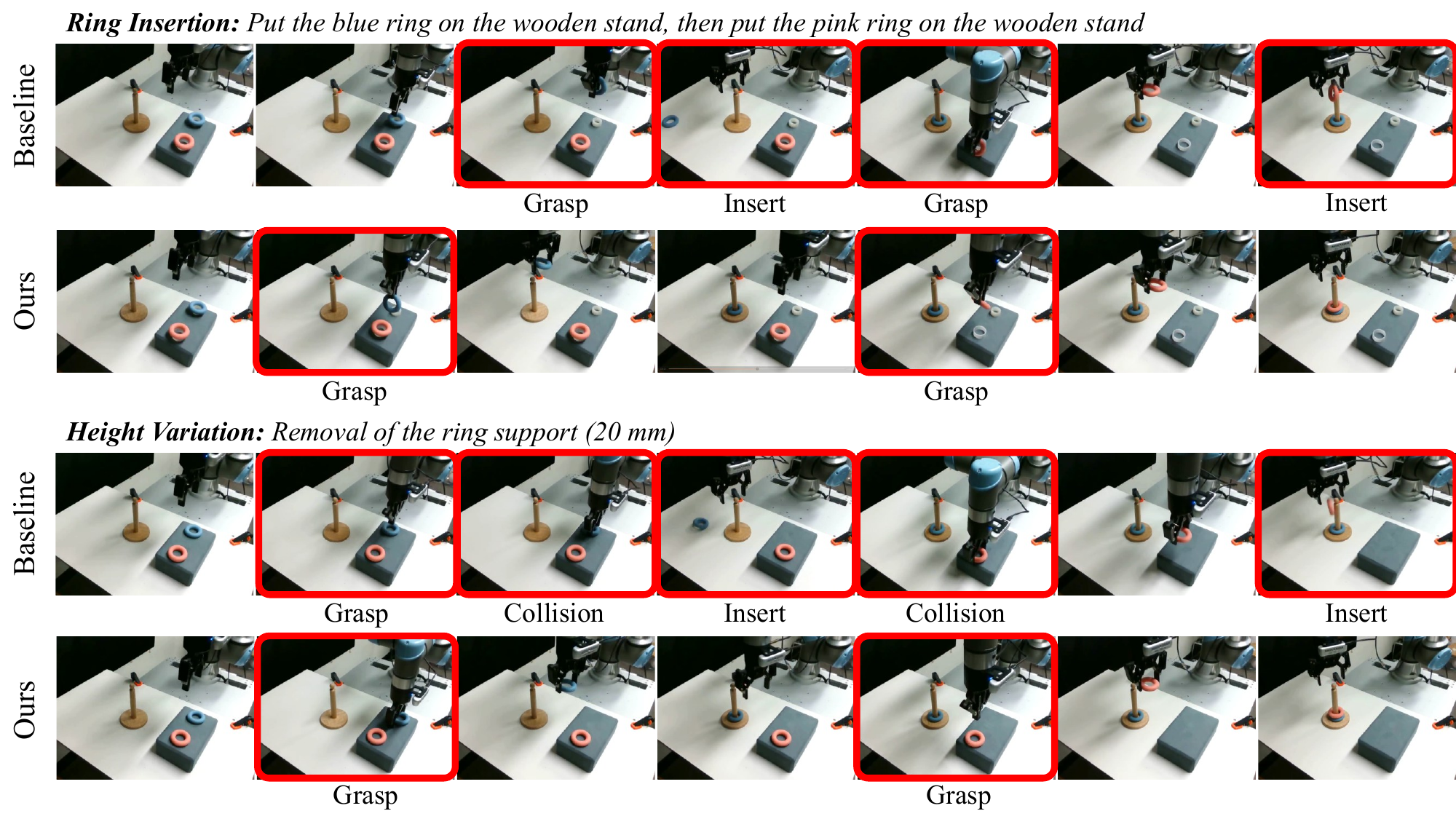}
    \caption{
   \textbf{Failure case visualization} on \textit{Ring Insertion} with \textbf{height variation}. 
    } 
    \label{figure:failurecase_ring}
\end{figure*}
}

\newcommand{\figRealsetting}{
\begin{figure*}[t!]
    \centering
    \includegraphics[width=1.0\textwidth, height=0.341\textwidth]{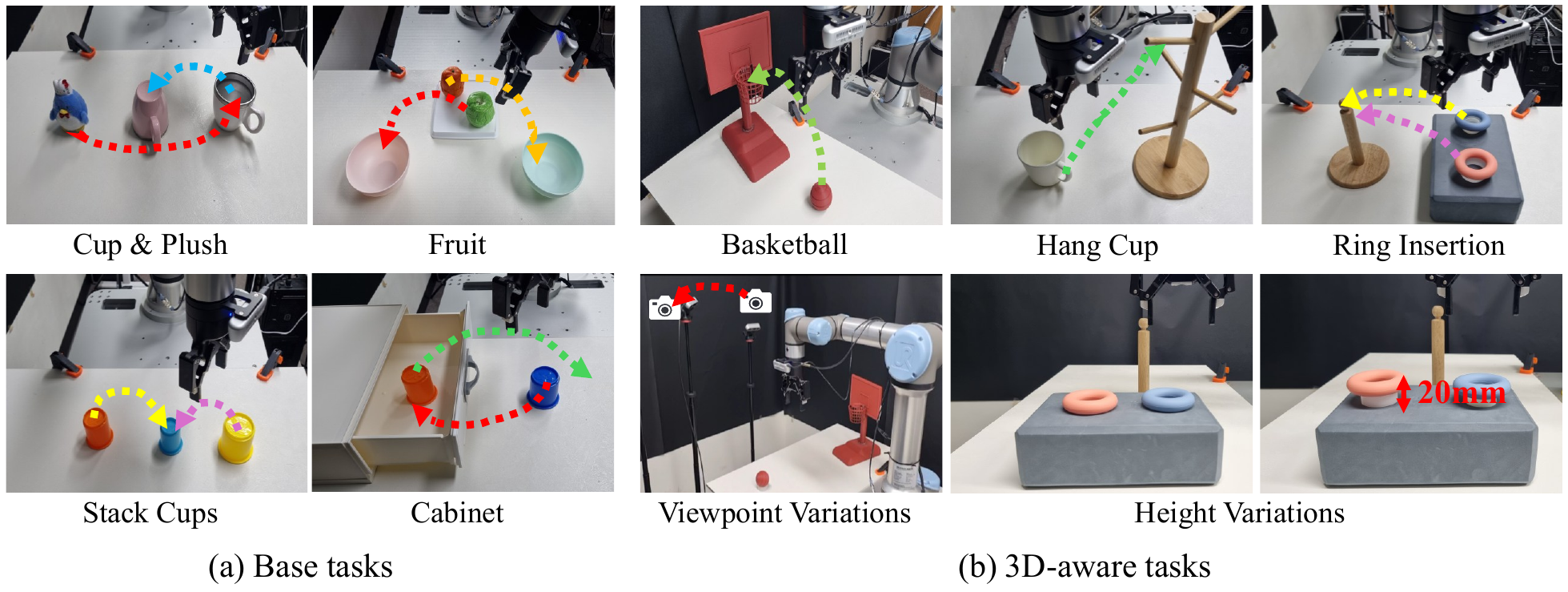}
    \caption{\textbf{Task visualization.} All policies are trained with only 40 demonstrations per task.}
    \label{figure:real_setting}
    \vspace{-2mm}
\end{figure*}
}

\newcommand{\figtaskwisefirst}{
\begin{figure*}[h!]
    \centering
    \includegraphics[width=1.0\textwidth, height=0.6\textwidth]{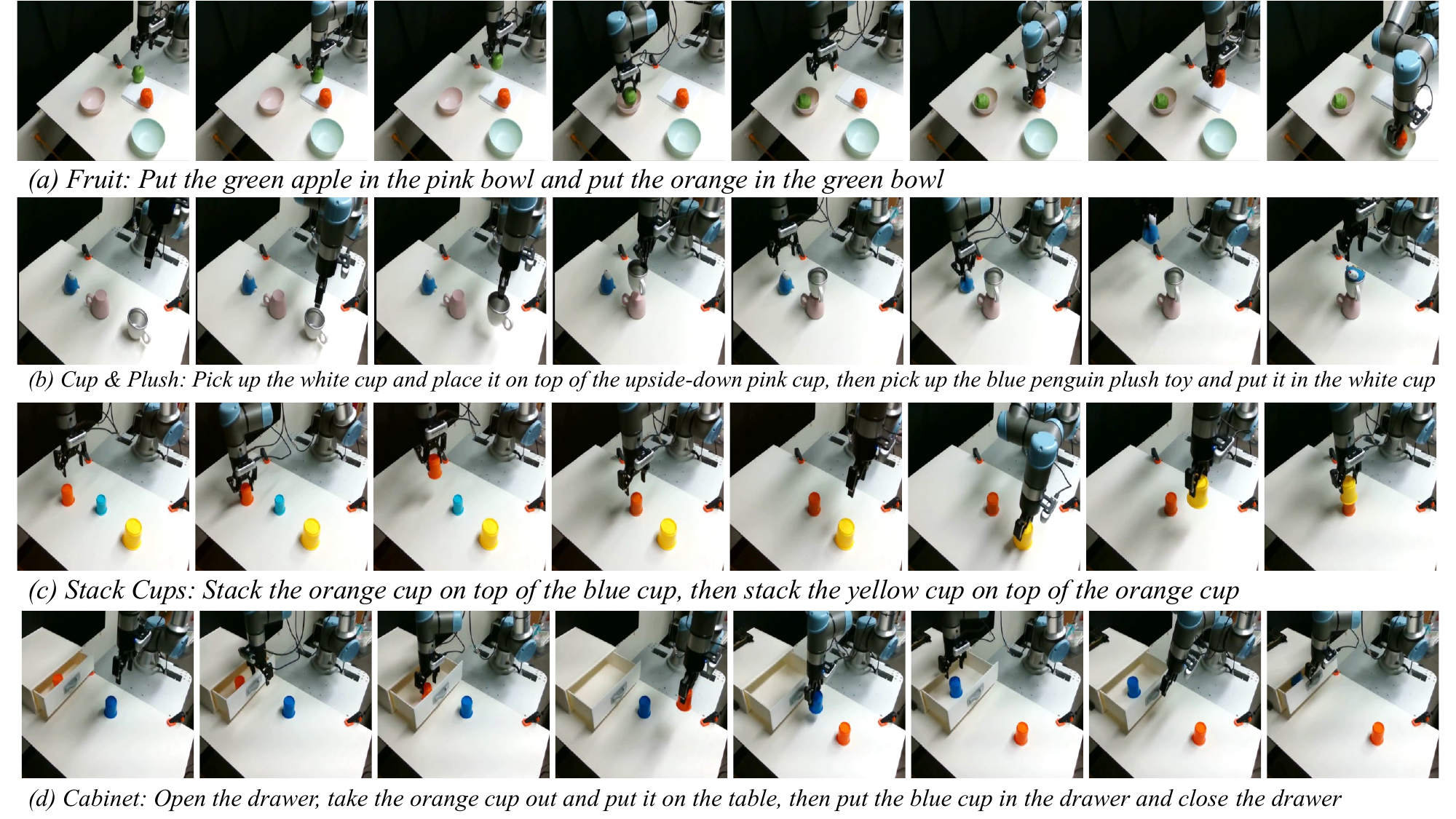}
   \caption{\textbf{Base task visualization.} Overview of the four long-horizon base tasks used in our real-world experiments: Fruit, Cup \& Plush, Stack Cups, and Cabinet.}
    \label{figure:taskwise_part1}
\end{figure*}
}

\newcommand{\figtaskwisesecond}{
\begin{figure*}[h!]
    \centering
    \includegraphics[width=1.0\textwidth, height=0.45\textwidth]{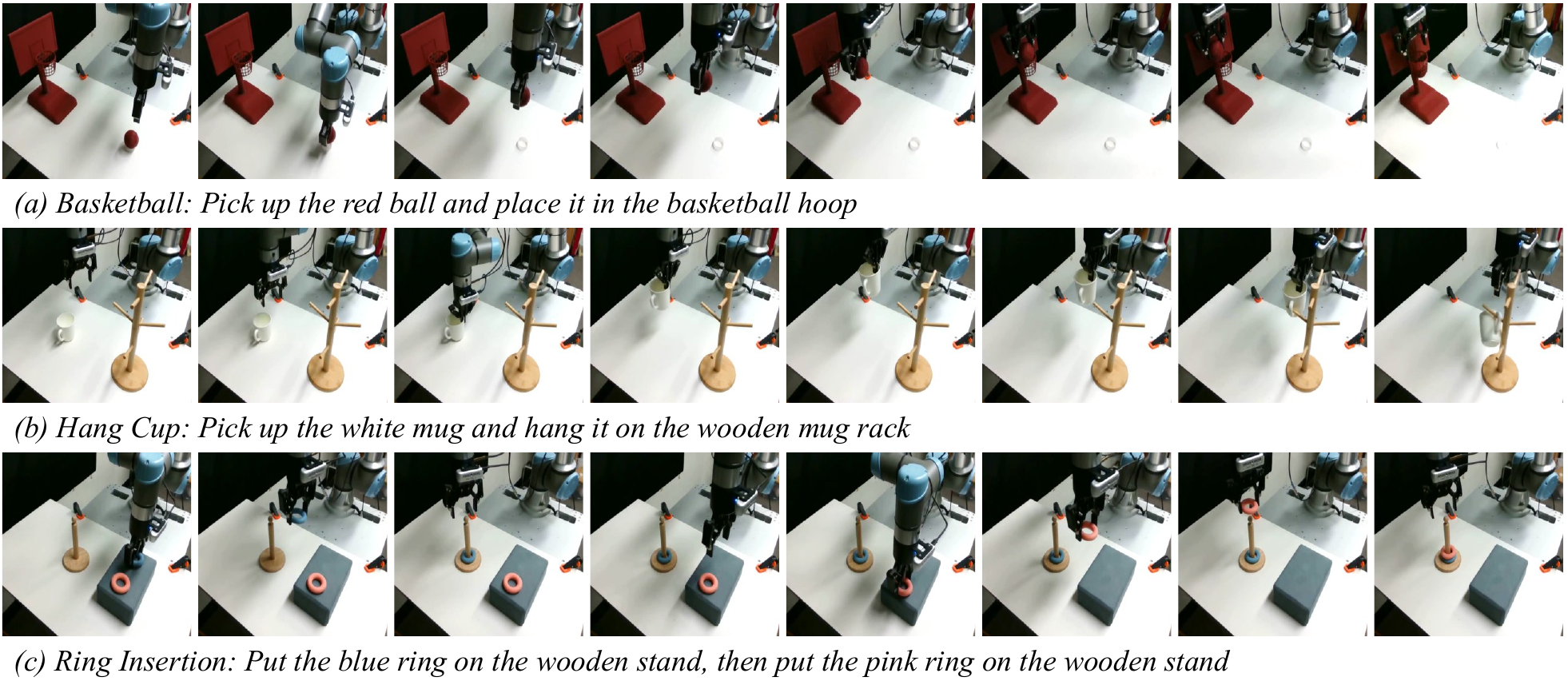}
   \caption{\textbf{3D-aware task visualization.} Overview of the three 3D-aware tasks used in our real-world experiments: Basketball, Ring Insertion, and Hang Cup.}
    \label{figure:taskwise_part2}
\end{figure*}
}

\newcommand{\figdeploysetting}{
\begin{figure*}[h!]
    \centering
    \includegraphics[width=0.7\textwidth, height=0.3\textwidth]{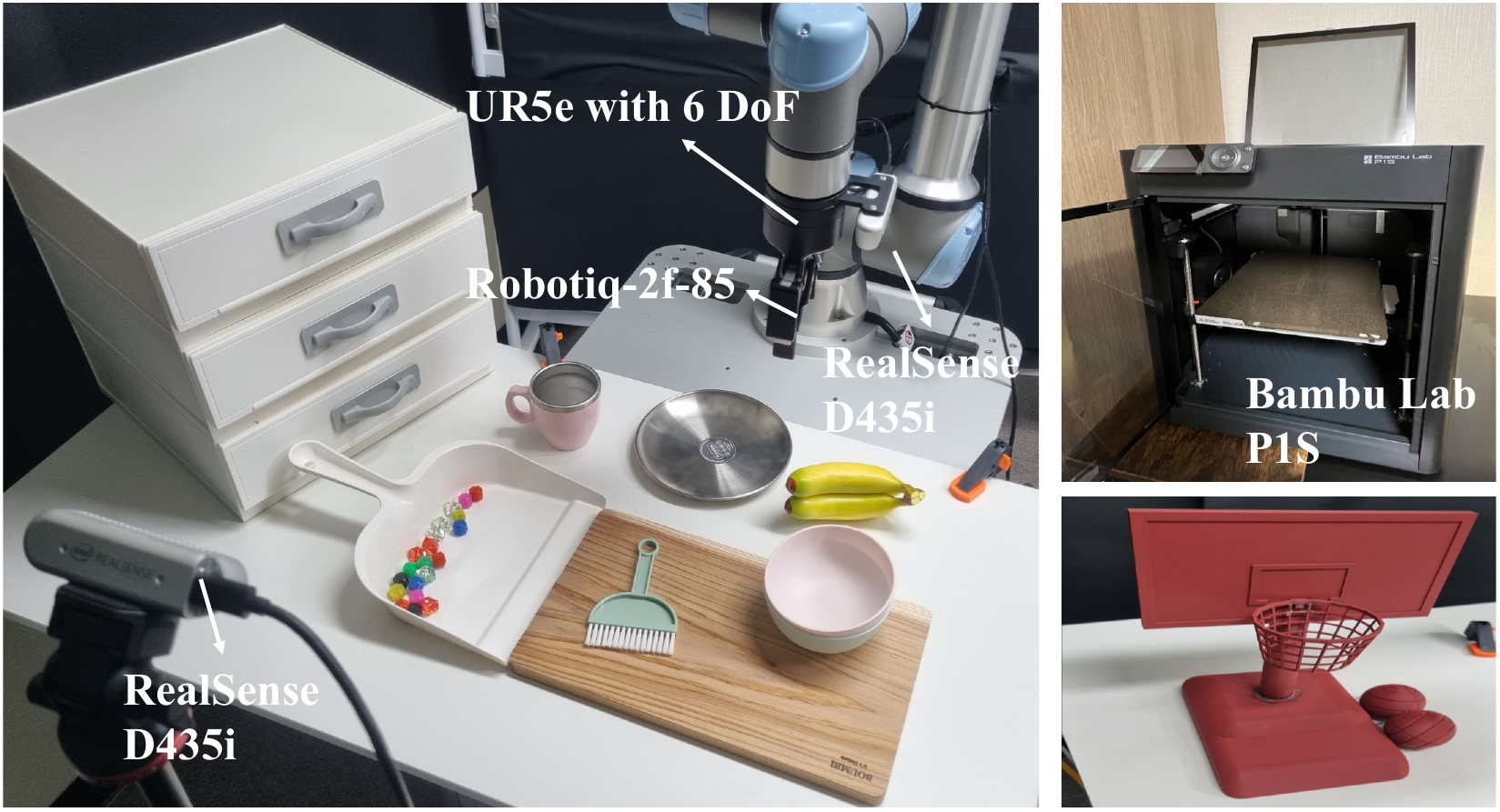}
    \caption{\textbf{Real-world deployment setup.}}
    \label{figure:real_deploy}
    \vspace{-4mm}
\end{figure*}
}

\newcommand{\tabLIBERO}{
\begin{table*}[t]
\setlength{\tabcolsep}{9.6pt}
\centering
{\footnotesize
\caption{
\textbf{Results on LIBERO Benchmark.} For Seer~\cite{seer}, we follow the original LIBERO-Long evaluation protocol, averaging the top-3 checkpoints over 20 rollouts per task.}
\label{table:LIBERO}
\begin{tabular}{ccccc|c}
\toprule
Method   & Spatial (\%) & Object (\%) & Goal (\%)  & Long (\%) & Avg. (\%) \\
\midrule
\multicolumn{1}{l}{Diffusion Policy \textit{(RSS'23)}} &        78.3 & 92.5 & 68.3 & 50.5 & 72.4 \\
\multicolumn{1}{l}{OpenVLA \textit{(CoRL'24)}}     &       84.7 & 88.4 & 79.2 & 53.7 & 76.5 \\
\multicolumn{1}{l}{$\pi_0$ \textit{(RSS'25)}}      &       96.8  &  98.8   &  95.8  & 85.2 & 94.2 \\
\multicolumn{1}{l}{CoT-VLA \textit{(CVPR'25)}}     &       87.5 & 91.6 & 87.6 & 69.0 & 83.9 \\
\multicolumn{1}{l}{$\pi_0$-FAST \textit{(RSS'25)}} &    96.4 & 96.8 & 88.6 & 60.2 & 85.5 \\
\multicolumn{1}{l}{SpatialVLA \textit{(RSS'25)}}   &  88.2 & 89.9 & 78.6 & 55.5 & 78.1 \\
\multicolumn{1}{l}{Octo \textit{(RSS'24)}}         &  78.9 & 85.7 & 84.6 & 51.1 & 75.1 \\
\multicolumn{1}{l}{UniVLA \textit{(RSS'25)}}       &  96.5 & 96.8 & 95.6 & 92.0 & 95.2 \\
\midrule
\multicolumn{1}{l}{Seer \textit{(ICLR'25)}}        &  84.2   &  94.8      &   89.8   & 80.4 & 87.3 \\
\rowcolor{gray!30}\multicolumn{1}{l}{\textbf{Seer + FlowVLA (Ours)}}&  93.5     &   96.5     &  91.8    & 87.8 & 92.4 \\
\multicolumn{1}{l}{$\pi_{0.5}$ \textit{(CoRL'25)}} &   98.8  &  98.2   &  98.0  & 92.4 & 96.8 \\
\rowcolor{gray!30}\multicolumn{1}{l}{\textbf{$\pi_{0.5}$ + FlowVLA (Ours)}} &   99.0     &   98.8     &  97.4     & 94.8 & 97.5 \\
\bottomrule
\end{tabular}
}
\end{table*}
}

\newcommand{\tabPLUS}{
\begin{table*}[t]
\setlength{\tabcolsep}{5.6pt}
\centering
{\footnotesize
\caption{
\textbf{Zero-shot evaluation on LIBERO-Plus.} Results demonstrate the robustness of policies trained on LIBERO benchmarks under unseen environmental perturbations.}
\label{table:LIBERO_plus}
\begin{tabular}{lccccccc|c}
\toprule
        & Camera & Robot & Language & Light & Background & Noise & Layout & Avg. (\%) \\
\midrule
OpenVLA & 0.8    & 3.5   &  23.0    & 8.1   &  34.8      &15.2   &28.5    & 15.6   \\
WorldVLA & 0.1   & 27.9  & 41.6     & 43.7  & 17.1       &10.9   & 38.0   & 25.0 \\
UniVLA  & 1.8    & 46.2  & 69.6     & 69.0  & 81.0       & 21.2  & 31.9   & 42.9 \\
\midrule
Seer    & 22.1   & 47.7  &  86.3    & 12.3  & 2.0        & 22.9 & 50.0    & 36.1  \\
\rowcolor{gray!30}\textbf{Seer + FlowVLA} & 24.6 & 54.6 & 87.0    &25.1   & 8.1        & 25.5 & 57.2    & 41.1  \\
$\pi_{0.5}$ & 21.4 & 39.1 & 68.4 & 86.7 & 83.3 & 36.1 & 84.6 & 57.5 \\
\rowcolor{gray!30}\textbf{$\pi_{0.5}$ + FlowVLA} & 24.1 & 39.9 & 68.8 & 90.4 & 87.1 & 37.7 & 84.7 & 59.2 \\

\bottomrule
\end{tabular}
}
\vspace{-3mm}
\end{table*}
}

\newcommand{\tabSupervisionSignal}{
\begin{wraptable}{r}{0.377\linewidth}
\vspace{-4mm}
\centering
\setlength{\tabcolsep}{0pt}
\caption{\textbf{Effect of supervision signal.}}
\label{table:supervision_signal}
{\footnotesize
\begin{tabular}{lc}
\toprule
Supervision Signal & Avg. (\%) \\
\midrule
\textit{None (Baseline)} & 80.4 \\
2D optical flow (CoTracker~\cite{karaev2024cotracker}) & 84.5 \\
Static 3D point cloud & 85.0 \\
\rowcolor{gray!30}
\textbf{Dynamic 3D flow (Ours)} & \textbf{87.8} \\
\bottomrule
\addlinespace[1pt]
\end{tabular}
}
\vspace{-5mm}
\end{wraptable}
}

\newcommand{\tabrealvariation}{
\begin{wraptable}{r}{0.37\linewidth}
\vspace{-4mm}
\centering
\setlength{\tabcolsep}{3.5pt}
\caption{\textbf{Test-time variation results.}}
\label{table:variation}
{\footnotesize
\begin{tabular}{lcc}
\toprule
Method & Viewpoint & Height \\
\midrule
Seer & 0 & 46.7\\
\rowcolor{gray!30}\textbf{Seer + FlowVLA} & 46.7 & 73.3\\
\bottomrule
\end{tabular}
}
\vspace{-4mm}
\end{wraptable}
}

\newcommand{\tabreappendixtopk}{
\begin{wraptable}{r}{0.4\linewidth}
\vspace{-4mm}
\centering
\caption{\textbf{Ablation on the number of Top-$K$.}}
\label{table:topk_ablation}
{\footnotesize
\begin{tabular}{cccc}
\toprule
Top-$K$ & 1024 & 2048 & 3072 \\
\midrule
Success (\%) & 86.0 & \textbf{87.8} & 85.2\\
\bottomrule
\end{tabular}
}
\vspace{-4mm}
\end{wraptable}
}

\newcommand{\tabreappendixvggt}{
\begin{wraptable}{r}{0.55\linewidth}
\centering
\setlength{\tabcolsep}{2.5pt}
\caption{\textbf{Ablation on reconstruction models for 3D flow supervision.}}
\label{table:vggt_ablation}
{\footnotesize
\begin{tabular}{ccccc}
\toprule
Method & Baseline & VGGT & SpatialTrackerV2 &VDPM \\
\midrule
Success (\%) & 80.4 & 86.8 & 88.0 & \textbf{87.8}\\
\bottomrule
\end{tabular}
}
\vspace{-4mm}
\end{wraptable}
}

\newcommand{\tabreappendixgaussiannoise}{
\begin{wraptable}{r}{0.4\linewidth}
\vspace{-4mm}
\centering
\setlength{\tabcolsep}{2.5pt}
\caption{\textbf{Noise sensitivity analysis.}}
\label{table:gaussian_noise}
{\footnotesize
\begin{tabular}{cccc | c}
\toprule
Noise ratio (\%) & 0 & 5 & 20 & Baseline \\
\midrule
Success (\%) & \textbf{87.8} & 84.7 & 84.0 & 80.4 \\
\bottomrule
\end{tabular}
}
\vspace{-4mm}
\end{wraptable}
}

\newcommand{\tabFigEfficiency}{
\begin{figure}[t]
\centering

\begin{minipage}[t]{0.5\linewidth}
\vspace{0pt}
\centering
\setlength{\tabcolsep}{4.9pt}
\captionof{table}{\textbf{Component analysis on LIBERO-Long.}}
\label{table:component_ablation}
{\footnotesize
\begin{tabular}{lc}
\toprule
Method & Avg. (\%) \\ \midrule
\rowcolor{gray!30}
\textbf{FlowVLA (Ours)} & \textbf{87.8} \\
\enspace $-$ w/o coordinate alignment & 86.3 \\
\enspace $-$ w/o reference-frame alignment & 85.2 \\
\enspace $-$ w/o magnitude-based selection & 84.0 \\
\enspace $-$ w/o TCP-level loss & 83.8 \\
\enspace $-$ w/o point-level loss & 84.8 \\
\midrule
\textit{None (Baseline)} & 80.4 \\
\bottomrule
\end{tabular}
}
\end{minipage}
\hfill
\begin{minipage}[t]{0.49\linewidth}
\vspace{0pt}
\centering
\captionof{figure}{\textbf{Analysis of training data efficiency and computational cost.}}
\includegraphics[width=1.03\linewidth, height=0.5\linewidth]{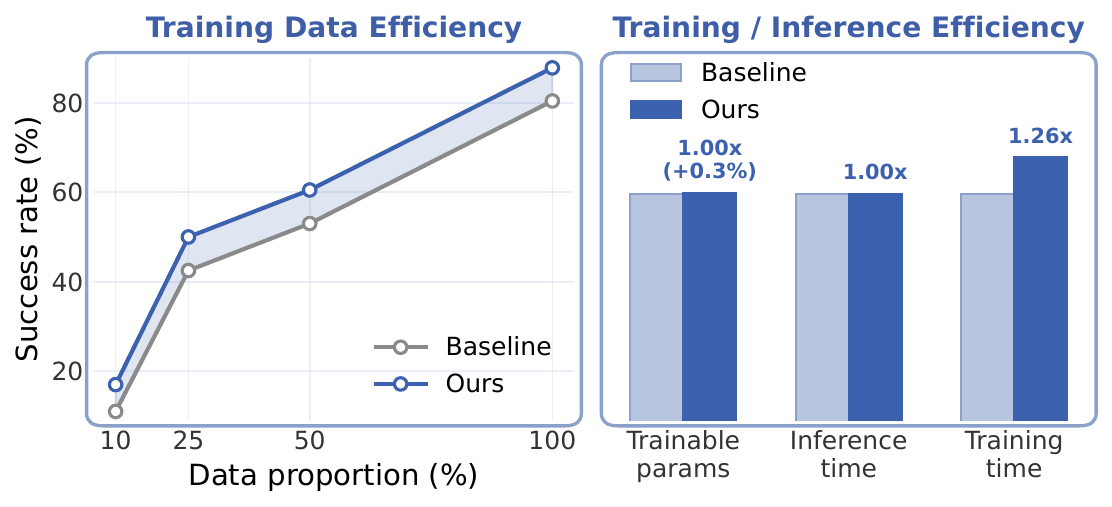}
\label{figure:training_inference}
\end{minipage}
\vspace{-9mm}
\end{figure}
}

\newcommand{\tabrealdeploy}{
\begin{table*}[h!]
\setlength{\tabcolsep}{4.5pt}
\centering
{\footnotesize
\caption{\textbf{Real-world results}. We evaluate each task over 15 trials and report success rates (\%).}
\label{table:realdeploy}
\begin{tabular}{lccccccc}
\toprule
        & \multicolumn{4}{c}{Base Tasks} 
        & \multicolumn{3}{c}{3D-aware Tasks} \\
\cmidrule(lr){2-5} \cmidrule(lr){6-8}
        & Cup \& Plush & Fruit & Stack Cups & Cabinet 
        & Basketball & Hang Cup & Ring Insertion \\
\midrule
Seer    &  60.0 & 66.7 & 73.3 & 60.0 & 60.0 & 46.7 & 66.7\\
\rowcolor{gray!30}
\textbf{Seer + FlowVLA} 
        & 73.3 & 73.3 & 93.3 & 73.3 & 80.0 & 86.7 & 73.3\\
\bottomrule
\end{tabular}
}
\end{table*}
}

\newcommand{\tabdeploydetail}{
\begin{table}[h]
\setlength{\tabcolsep}{15pt}
\centering
\caption{\textbf{Experimental protocol and hyperparameter configuration for Seer~\cite{seer}.}}
\label{tab:protocol_hparams}
\begin{tabular}{ccc}
\toprule
\textbf{Category} & \textbf{Simulation} &\textbf{Real-world} \\
\midrule
Evaluation Trials & 20 per task & 15 per task \\
Training Demonstrations & 50 per task & 40 per task \\
Test-time Randomization & - & position perturbations \\
Batch Size & 512 & 512 \\
Learning Rate & 1e-3 & 1e-3\\
LR Schedule & cosine & cosine\\
Prediction Length $k$ & 3 & 3\\
Historical Length $L$ & 7 & 7 \\   
Selected Points $K$ & 2048 & 2048 \\
$\lambda_{\mathrm{point}}$ & 0.005 & 0.05 \\
$\lambda_{\mathrm{tcp}}$ & 0.005 & 0.05\\
\bottomrule
\end{tabular}
\end{table}
}

\newcommand{\tabsupplecomponent}{
\begin{wraptable}{r}{0.43\linewidth}
\vspace{-4.3mm}
\centering
\setlength{\tabcolsep}{3pt}
\caption{\textbf{Component analysis under viewpoint variation.}}
\label{tab:component_viewpoint}
{\footnotesize
\begin{tabular}{lc}
\toprule
Method & Success \\
\midrule
\rowcolor{gray!30}
\textbf{FlowVLA} & \textbf{7/15} \\
\enspace $-$ w/o coordinate alignment & 4/15 \\
\enspace $-$ w/o magnitude-based selection & 0/15 \\
\enspace $-$ w/o TCP-level loss & 2/15 \\
\enspace $-$ w/o point-level loss & 4/15 \\
\bottomrule
\end{tabular}
}
\vspace{-5mm}
\end{wraptable}
}

\newcommand{\tabrealpi}{
\begin{wraptable}{r}{0.53\linewidth}
\vspace{-4mm}
\centering
\setlength{\tabcolsep}{3.5pt}
\caption{\textbf{Real-world results with $\pi_{0.5}$.}
Evaluation on 3D-aware tasks and test-time dynamic-background variation.}
\label{table:realpi}
{\footnotesize
\begin{tabular}{lccc}
\toprule
Method & Basketball & Hang Cup & Background\\
\midrule
$\pi_{0.5}$ & 10/15 & 9/15 & 5/15 \\
\rowcolor{gray!30}\textbf{$\pi_{0.5}$ + FlowVLA} & 13/15 & 14/15 & 8/15\\
\bottomrule
\end{tabular}
}
\vspace{-4mm}
\end{wraptable}
}

\begin{abstract}
Recent advances in 3D reconstruction enable the extraction of temporally consistent 3D motion from RGB observations; however, current Vision-Language-Action (VLA) models still predominantly rely on 2D image-text priors.
Therefore, they do not accurately represent the fundamental dynamics of the 3D world associated with physical actions.
This over-reliance on static 2D visual cues limits the capacity of VLA models in achieving robust robot manipulation.
To address these limitations, we introduce \textbf{FlowVLA}, a reconstruction-guided, pluggable 3D flow learning framework aimed at empowering VLA models with an inherent understanding of action-relevant dynamics. 
Instead of requiring explicit 3D sensor inputs or altering the inference-time architecture, our method is designed to utilize temporally aligned 3D flow only during training.
Our method allows the model to learn the dynamic evolution of the physical scene while preserving the computational efficiency of conventional 2D VLA backbones.
To bridge scene-level dynamics with robotic manipulation, we propose a modular 3D flow learning framework defined by a dual-level flow supervision design.
Specifically, point-level flow guides the model to forecast the future displacements of task-critical 3D points, while TCP-level flow aggregates these motions to align with the robot’s Tool-Center-Point (TCP) trajectories.
This dual-level flow supervision encourages the model to focus on semantically meaningful dynamic regions, effectively coupling scene-level motion with precise control signals. 
Extensive evaluations across simulation and real-world experiments demonstrate that FlowVLA consistently improves manipulation performance, zero-shot robustness, and optimizes data efficiency. 
FlowVLA adds only lightweight auxiliary heads during training and discards them at inference.
Therefore, FlowVLA introduces no additional 3D sensor inputs, architectural modifications, or inference-time computation.

\end{abstract}

\section{INTRODUCTION}
Reliable robotic manipulation should consider interaction within a dynamic, three-dimensional physical world.
This dynamic interaction is addressed through the emergence of VLA models as a strong paradigm, transferring the rich semantic knowledge of pre-trained LLMs and VLMs to physical robot controls~\cite{rt1, rt2, black2024pi0, pi05, shukor2025smolvla}.
By integrating large-scale multimodal backbones with an action decoder, recent VLAs have achieved remarkable performance across diverse robotic tasks, including complex scene reasoning~\cite{ecot, cotvla}, cross-embodiment generalization~\cite{oxe, pi05, team2025gemini}, and language-instructed manipulation~\cite{kim2024openvla, black2024pi0, team2024octo}.

Despite their impressive capabilities, most existing VLA models remain constrained by their dependence on multimodal representations derived from 2D image-text pretraining.
Although these 2D representations provide rich semantic priors, they are not inherently designed to capture action-relevant 3D dynamics required for real-world robot manipulation.
Consequently, action policies often overfit to static visual cues---such as backgrounds, static camera viewpoints, or absolute global geometry---rather than prioritizing action-relevant 3D interactions and dynamics.
To mitigate these limitations, recent approaches~\cite{qu2025spatialvla, li2025bridgevla, li2026pointvla, 3DVLA} have explored 3D-aware VLAs that integrate explicit 3D signals, including depth maps or point clouds.
Although these approaches improve physical manipulation by leveraging spatial understanding, they also present new challenges.
In their approaches, auxiliary 3D sensors are required, and costly retraining on paired 3D-action datasets is demanded.
More crucially, these approaches emphasize the extraction of static 3D geometry from single observations; as a result, they still overlook the dynamic evolution of the scene that is directly tied to physical manipulation.

To address these challenges, we propose \textbf{FlowVLA}, a reconstruction-guided, pluggable 3D flow learning framework that injects action-relevant 3D motion cues into VLA representations.
Our core insight is to leverage 3D flow exclusively as a privileged signal during training to inject manipulation-relevant 3D dynamics directly into the model's latent representations. 
Instead of providing static snapshots, 3D flow provides direct supervision on how the physical environment moves and evolves under robot actions.
Therefore, we employ 3D flow not as an input modality, but as a training-only module, while preserving the architecture of VLAs during inference.
To achieve this, we derive temporally aligned 3D point trajectories from consecutive frames~\cite{vdpm}.
By computing the positional differences between adjacent frames, we formulate 3D flow that highlights regions of real physical movement within the scene.
To effectively bring this spatial motion into the learning process, we establish a dual-level scene-to-control 3D flow learning framework comprising scene-level and control-level modules.
The scene-level module captures dense 3D scene dynamics by supervising the future displacements of motion-salient 3D points, while the control-level module distills these dense motions into TCP-aligned flow that reflects the robot end-effector trajectory.
By jointly optimizing these two modules, our framework bridges holistic environmental changes in 3D space with control-relevant robot motion.

We comprehensively validate FlowVLA through extensive evaluations on standard benchmarks, such as LIBERO and LIBERO-Plus, and real-world robot experiments.
Empirical results demonstrate that our 3D sensor-free framework improves diverse VLA models with zero inference overhead.
By improving task success rates in both simulated and real-world environments, the proposed approach highlights the potential of FlowVLA as a practical framework for 3D-aware policy learning.

\section{RELATED WORK}
\subsection{Vision-Language-Action Models}
VLAs integrate visual observations, proprioceptive states, and language instructions to generate robot actions. 
The viability of language-model-based sequence modeling for robot control was established by early studies~\cite{saycan,gato,palme},
leading to RT-1~\cite{rt1}.
RT-2~\cite{rt2} further showed that the semantic knowledge acquired from internet-scale pretraining can transfer effectively to robot manipulation.
Recent advancements have concentrated on scaling, generalization, and expressive action generation. Octo~\cite{team2024octo} and OpenVLA~\cite{kim2024openvla} utilize large-scale robot datasets~\cite{oxe, bridgedata} to improve adaptation to novel environments. 
Other works extend VLA models with continuous action generation via flow matching or diffusion~\cite{black2024pi0, pi05, wen2025diffusionvla}, internal reasoning mechanisms~\cite{ecot, cotvla}, and world modeling~\cite{seer, cen2025worldvla}.
Even with these advances, the current VLA paradigm still relies on representations inherited from 2D image-text pretraining.
This makes existing models highly effective at semantic grounding, but leaves action-relevant 3D dynamics insufficiently modeled~\cite{3DVLA, LEO, chen2023polarnet}.

\subsection{3D-Aware VLA Models}
To improve spatial grounding in VLA models, recent works have incorporated explicit 3D information into their pipelines. 
Models such as LLaVA-3D~\cite{llava3d}, SpatialVLA~\cite{qu2025spatialvla}, and DM0~\cite{yu2026dm0} exploit 3D representations to enhance spatial understanding, whereas 3DVLA~\cite{3DVLA} injects 3D features into the VLM embedding space for improved reasoning capabilities.
It is demonstrated by these methods that useful physical grounding for robot manipulation can be provided by 3D structures.
However, incorporating 3D information into 2D-pretrained backbones poses significant challenges. Dedicated alignment mechanisms are required due to the domain gap between pixel-based and geometric representations~\cite{li2026pointvla, zhou2025cross, li2025bridgevla, chen2023bridging}. 
Furthermore, high-quality 3D-action paired datasets are remarkably scarce compared to large-scale 2D corpora, making the computational cost of retraining prohibitively expensive~\cite{3DVLA, xu2024pointllm}.
Several recent methods therefore explore more efficient ways to integrate 3D information. PointVLA~\cite{li2026pointvla} introduces lightweight modules for injecting 3D information without retraining the entire backbone, while GeoVLA~\cite{sun2025geovla} employs a parallel 3D embedding network to preserve multimodal alignment. BridgeVLA~\cite{li2025bridgevla} instead projects 3D point clouds into multiple 2D views so that they can be processed within a unified representation space.
Despite their effectiveness, these approaches typically rely on 3D sensor-derived inputs or 3D representations, and primarily emphasize static 3D geometry while providing limited supervision on how motion-salient regions physically evolve under robot actions.
In contrast, \textbf{FlowVLA} uses reconstruction-derived 3D flow as a training-only privileged signal, encouraging existing 2D VLA policies to encode action-relevant 3D motion cues without the need for auxiliary 3D sensors or additional inference-time computation.

\section{METHOD}
As shown in Fig.~\ref{figure:overall}, we present \textbf{FlowVLA}, a reconstruction-guided 3D flow learning framework that injects action-relevant 3D dynamics into VLA models through training-only supervision.
Temporally aligned 3D point trajectories are derived from RGB sequences by our FlowVLA using a feed-forward dynamic reconstruction model~\cite{vdpm}.
Then, it converts 3D point trajectories into future 3D flow targets.
These targets supervise the VLA with a dual-level scene-to-control objective: the action latent is encouraged by point-level flow to encode dense 3D scene motion, while motion cues are distilled into a control-oriented representation aligned with end-effector displacement by TCP-level flow.

\subsection{Preliminaries}\label{preliminaries}
\paragraph{VDPM~\cite{vdpm}.}
We adopt VDPM, a model intended to recover temporally consistent 3D point trajectories from RGB observations, to obtain a 3D motion signal without any other sensors.
The core advantage of VDPM is its ability to consistently track points defined in a reference frame over time.
Specifically, given an observation $o_{t_{\mathrm{ref}}}$ at reference time step $t_{\mathrm{ref}}$, VDPM represents the 3D location of each point on the reference pixel grid $u \in \{0,\dots,H-1\} \times \{0,\dots,W-1\}$ at an arbitrary time step $t^\prime$ as $P_{t_{\mathrm{ref}}}(t^\prime, \pi)\in\mathbb{R}^{H\times W\times 3}$, where $\pi$ denotes the coordinate system in which the 3D points are expressed.
Thus, $P_{t_\mathrm{ref}}(t^\prime, \pi)(u)$ is the 3D position of the point corresponding to pixel $u$ in the reference frame, evaluated at time step $t^\prime$.
For fixed $t_{\mathrm{ref}}$ and $\pi$, the 3D trajectory of a reference point over a sequence is defined as $\{P_{t_{\mathrm{ref}}}(t^\prime, \pi)(u)\}_{t^\prime=0}^{T-1}$.

\paragraph{Coordinate Alignment.}
We set the first frame as the reference frame, i.e., $t_{\mathrm{ref}}=0$, and define the corresponding reference camera coordinate as $\pi_0$.
VDPM initially generates reference-anchored point trajectories within this reference camera coordinate, denoted as $P_0(t^\prime,\pi_0)(u)$.
However, robot motion, including TCP displacement, is defined in the robot base coordinate.
Before constructing flow targets, all 3D points are expressed in the robot base coordinate $\pi_{\mathrm{base}}$ to ensure that the reconstruction-derived point trajectories are directly comparable with robot motion.
For each task, we compute a rigid transform 
$T_{\mathrm{base}\leftarrow0}=[R_{\mathrm{base}\leftarrow0} \mid \mathbf{t}_{\mathrm{base}\leftarrow0}]\in SE(3)$
from the reference camera frame to the robot base frame.
Subsequently, we proceed to transform each reference-anchored 3D point as follows:
\begin{align}
    P_0(t^\prime,\pi_{\mathrm{base}})(u) = R_{\mathrm{base}\leftarrow 0}P_0(t^\prime,\pi_0)(u) + \mathbf{t}_{\mathrm{base}\leftarrow 0}.
\end{align}
For a trajectory of length $N$, we define the 3D point flow at time $t^\prime$ in the robot base coordinate as:
\begin{align}
    \Delta P_0(t^\prime,u) = P_0(t^\prime+1,\pi_{\mathrm{base}})(u) - P_0(t^\prime,\pi_{\mathrm{base}})(u), \quad t^\prime = 0,\dots,N-2.
    \label{eq:reference_delta}
\end{align}
Here, $\Delta P_0(t^\prime,u)\in\mathbb{R}^{3}$ is referred to as the 3D displacement of reference point $u$ from $t^\prime$ to $t^\prime+1$ in the robot base coordinate.
This alignment assures that the resulting 3D point flows are defined in the same coordinate system as the TCP displacements used in the TCP-level objective.

\figOverall

\subsection{Reconstruction-Guided 3D Flow Construction}\label{sec:3dflow_extract}
Given visual observations $o_{t-L+1:t}$, a language instruction $l$, and proprioceptive states $s_{t-L+1:t}$, where $L$ denotes the historical observation length and $k$ denotes the future prediction horizon, the VLA backbone generates action latent representations $z_{t:t+k-1}$ as follows:
\begin{align}
    z_{t:t+k-1} &= f_{\theta}\!\left(f_v(o_{t-L+1:t}), f_l(l), f_s(s_{t-L+1:t})\right), \label{eq:latent} \\
    a_{t:t+k-1} &= p_{\theta}\!\left(z_{t:t+k-1}\right), \label{eq:action}
\end{align}
where $f_{\theta}$ and $p_{\theta}$ denote the VLA backbone and action policy, respectively, and $f_v$, $f_l$, and $f_s$ act as modality-specific encoders.
While rich semantic priors are provided by these 2D-centric representations, they often struggle to capture the underlying 3D dynamics essential for complex manipulation. 
The reference-anchored 3D flow defined in Eq.~\ref{eq:reference_delta} is utilized as a reconstruction-derived, training-only supervision signal to bridge this gap.

\paragraph{Magnitude-based Selection.} Since not all points contribute equally to action-relevant dynamics, we prioritize the selection of dynamically informative points according to their motion magnitude. 
This filtering directs the model's attention towards dynamically informative regions rather than static backgrounds or reconstruction artifacts. 
Specifically, for each point $u$ on the reference-image pixel grid, we compute its instantaneous flow magnitude at time $\tau$ within an observation window of length $L+k$ using the formula $m_{\tau}(u)=\left\|\Delta P_0(\tau,u)\right\|_2$, where $\tau$ ranges from $t-L+1$ to $t+k-1$.
The average flow magnitude over the sequence window is calculated as follows: 
\begin{align}
    \bar{m}_t(u) = \frac{1}{L+k-1}\sum_{\tau=t-L+1}^{t+k-1} m_{\tau}(u).
\end{align}
By selecting the top-$K$ points that exhibit the highest $\bar{m}_t(u)$ values, we construct a keypoint set \(\mathcal{U}_t = \operatorname{TopK}\big(\{\bar{m}_t(u)\}_{u}, K\big)\), which identifies regions with dominant 3D motion.
For each selected point $u \in \mathcal{U}_t$, the future 3D flow target at prediction step $j$ is defined as:
\begin{align}
    \Delta P_0^{(j)}(t,u):=\Delta P_0(t+j,u)=P_0(t+j+1,\pi_\mathrm{base})(u) - P_0(t+j,\pi_\mathrm{base})(u),
\end{align}
where $j\in\{0,\dots,k-1\}$ and $t+k \le N-1$. 
These selected points and their corresponding future displacements serve as ground-truth supervision signals for the dual-level flow learning objective, guiding the VLA action latent to encode the 3D evolution of the workspace.

\subsection{Dual-Level 3D Flow Learning}
\paragraph{Point-Level Flow. } 
The point-level flow branch facilitates the action latent representation in effectively capturing the future motion of dynamically informative 3D points.
With the selected keypoint set $\mathcal{U}_t$ and the future flow targets $\Delta P_0^{(j)}(t,u)$ from Sec.~\ref{sec:3dflow_extract}, the action latent token $z_{t+j}$ is first mapped to a higher-dimensional space and broadcast across the 
$K=|\mathcal{U}_t|$ points as follows:
\begin{align}
    \mathbf{h}_t^{(j)} = \operatorname{Broadcast}\big(g_{\phi}(z_{t+j}), K\big)\in\mathbb{R}^{K\times d_h}, \quad j\in\{0,\dots,k-1\}.
\end{align}
Here, $g_{\phi}$ is a 2-layer MLP and $d_h$ represents the latent dimension. 
To provide spatial context during training, we collect the 3D coordinates of the selected points at time step $t+j$ as:
\begin{align}
    \mathbf{P}_t^{(j)}=\big[P_0(t+j,\pi_\mathrm{base})(u)\big]_{u\in\mathcal{U}_t}\in\mathbb{R}^{K\times 3}.
\end{align}
We fuse the broadcast action feature with the selected 3D coordinates to obtain
$\mathbf{F}_t^{(j)}=q_{\phi}([\mathbf{h}_t^{(j)};\mathbf{P}_t^{(j)}])$,
where $q_{\phi}$ is a point-wise feature extractor.
The flow head $h_{\mathrm{flow}}$ predicts the future displacement of each selected point as
$\Delta \hat{\mathbf{P}}_0^{(j)}(t)=h_{\mathrm{flow}}(\mathbf{F}_t^{(j)})\in\mathbb{R}^{K\times 3}$.
These point-wise predictions are supervised by the point-level flow loss:
\begin{align}
    \mathcal{L}_{\mathrm{point}}=\frac{1}{kK}\sum_{j=0}^{k-1}\sum_{u\in\mathcal{U}_t}\rho\!\left(\Delta \hat{P}_0^{(j)}(t,u),\Delta P_0^{(j)}(t,u)\right),
\end{align}
where $\rho(\cdot,\cdot)$ denotes the Smooth L1 loss. 
This objective encourages the action latent to reflect scene-level 3D dynamics by predicting the future motion of selected points.

\paragraph{TCP-Level Flow.} 
While point-level supervision grounds the model in scene dynamics, not all motions are equally relevant to the control task. 
We aggregate the predicted point flows into a TCP flow estimate to focus on control-relevant dynamics. 
Importance weights $\mathbf{r}_t^{(j)}$ are predicted through a softmax-normalized weighting head $h_{\mathrm{weight}}$ using the fused representation $\mathbf{F}_t^{(j)}$:
\begin{align}
\mathbf{r}_t^{(j)}=\operatorname{Softmax}_{u}\!\left(h_{\mathrm{weight}}\!\left(\mathbf{F}_t^{(j)}\right)\right)\in\mathbb{R}^{K}. 
\end{align}
The predicted TCP-level flow is then computed as the weighted sum of point-wise displacements:
\begin{align}
    \Delta \hat{\mathbf{x}}_{\mathrm{tcp}}^{(j)}(t)= \sum_{u\in\mathcal{U}_t}\mathbf{r}_t^{(j)}(u)\,\Delta \hat{P}_0^{(j)}(t,u)\in \mathbb{R}^{3}.
\end{align}
The TCP-level flow loss is then defined as:
\begin{align}
    \mathcal{L}_{\mathrm{tcp}}=\frac{1}{k}\sum_{j=0}^{k-1}\rho\!\left(
    \Delta \hat{\mathbf{x}}_{\mathrm{tcp}}^{(j)}(t),\Delta \mathbf{x}_{\mathrm{tcp}}^{(j)}(t)\right),
\end{align}
where $\Delta \mathbf{x}_{\mathrm{tcp}}^{(j)}(t)=\mathbf{x}_{\mathrm{tcp}}(t+j+1)-\mathbf{x}_{\mathrm{tcp}}(t+j)$ is the ground-truth TCP flow.
This scene-to-control distillation encourages the action latent to capture control-relevant dynamics in addition to scene-level motion.
The final objective combines the action loss with the auxiliary flow losses:
\begin{align}
    \mathcal{L}_{\mathrm{total}}=\mathcal{L}_{\mathrm{action}}+ \lambda_{\mathrm{point}}\mathcal{L}_{\mathrm{point}}
    +\lambda_{\mathrm{tcp}}\mathcal{L}_{\mathrm{tcp}},
\end{align}
where $\lambda_{\mathrm{point}}$ and $\lambda_{\mathrm{tcp}}$ are the loss weights. 
Critically, all auxiliary heads and 3D geometric inputs are discarded at inference. 
The policy operates through its original 2D-based pathway, benefiting from 3D-aware representation with zero inference-time overhead.

\section{SIMULATION EXPERIMENTS}
\paragraph{Settings.}
The proposed 3D flow supervision is evaluated across heterogeneous policy designs to confirm the generalization ability of FlowVLA.
We instantiate FlowVLA on two representative VLAs with substantially different action-generation policies: Seer~\cite{seer} and $\pi_{0.5}$~\cite{pi05}. 
Seer adopts a two-layer MLP-based policy that reconstructs future observations and predicts actions, whereas $\pi_{0.5}$ employs a PaliGemma-based~\cite{beyer2024paligemma} VLM backbone with a flow-matching action policy. For $\pi_{0.5}$, we follow the official fine-tuning recipe and perform LoRA adaptation~\cite{hu2022lora}. 
For both backbones, we attach 3D flow objectives only during training while preserving their model architectures and inference pathways.
Further details on the simulation settings and datasets are provided in Appendix~\ref{appendix_simulation_setting}.

\subsection{Simulation Results}
\paragraph{LIBERO.} Tab.~\ref{table:LIBERO} summarizes the results on the LIBERO benchmark. 
Our scene-to-control 3D flow supervision yields consistent improvements on recent VLA models, with particularly large gains in long-horizon manipulation.
It is notable that FlowVLA achieves particularly large gains on \textbf{LIBERO-Long}, which requires sustained spatial-temporal reasoning over extended horizons. On Seer~\cite{seer}, our method improves the success rate from 80.4\% to 87.8\%, a substantial 7.4\%p increase. Even on the state-of-the-art $\pi_{0.5}$, FlowVLA improves the success rate from 92.4\% to 94.8\%.
Importantly, these performance gains are achieved with \textit{zero inference-time overhead}, as the auxiliary flow heads are entirely decoupled after training.

\tabLIBERO

\tabPLUS
\paragraph{LIBERO-Plus.} 
We conduct an additional evaluation of zero-shot robustness on LIBERO-Plus, and report the results in Tab.~\ref{table:LIBERO_plus}. 
The average success rate is improved from 36.1\% to 41.1\% on Seer~\cite{seer} model. The performance improvements are particularly noticeable for perturbation types such as \textit{Camera}, \textit{Robot}, \textit{Light}, \textit{Noise}, and \textit{Layout}. This suggests that reconstruction-derived 3D flow supervision guides the policy to focus on action-relevant 3D dynamics, rather than overfitting to superficial visual appearances. 
Notably, we also observe a modest gain under the \textit{Language} perturbation, indicating that the proposed method enhances dynamic motion grounding without compromising the semantic priors acquired from image-text pretraining.

\tabFigEfficiency

\subsection{Ablation Studies}
\paragraph{Component Analysis.}
We verify the contribution of each design choice through a component ablation study performed on \textbf{LIBERO-Long} (Tab.~\ref{table:component_ablation}).
\textbf{\ding{182} Coordinate alignment:}
Aligning 3D point trajectories across different camera coordinate systems improves performance, indicating that robot-aligned flow targets are important for consistent supervision.
\textbf{\ding{183} Reference-frame alignment:}
Using reference-anchored, temporally trackable 3D points further improves performance, supporting their effectiveness for defining reliable 3D flow targets.
\textbf{\ding{184} Magnitude-based selection:}
Replacing magnitude-based point selection with \textit{random-$K$ selection} reduces performance, indicating that directing supervision towards dynamic, motion-salient regions is important for efficient action-relevant learning.
\textbf{\ding{185} Dual-level flow supervision:}
Removing either \textit{TCP-level flow} or \textit{point-level flow} degrades performance, demonstrating that FlowVLA benefits from both scene-level motion grounding and end-effector-aligned control supervision.
Additional qualitative visualizations of the magnitude-based point selection and TCP-level aggregation weights are provided in Appendix~\ref{appendix:visualization}.

\paragraph{Data Efficiency and Computational Cost.} In Fig.~\ref{figure:training_inference}, we analyze FlowVLA in terms of data efficiency and computational cost. 
FlowVLA consistently outperforms the baseline across all data regimes when trained with 10\%, 25\%, and 50\% of the full dataset. 
This suggests that utilizing training-only 3D flow supervision provides a useful inductive bias under limited data, allowing for less reliance on superficial visual correlations and better capture of control-relevant scene dynamics by the policy. 
It is also evident that the computational overhead is kept at a manageable level: the proposed flow heads increase the total number of parameters by only 0.34\% and add no inference-time computation. The only cost occurs during training, where total training time increases by approximately 1.26$\times$. This trade-off is favorable given the consistent gains across evaluated tasks and backbones.
    
\tabSupervisionSignal
\paragraph{Effect of Supervision Signal.}
We compare different supervision targets under the same training and evaluation protocol.
Tab.~\ref{table:supervision_signal} shows that dynamic 3D flow supervision outperforms both 2D optical flow and static 3D point supervision.
In contrast to 2D optical flow, 3D flow captures motion in physical space instead of image-plane displacement.
This allows it to align more closely with scene-level dynamics and end-effector motion.
Unlike static 3D point supervision, 3D flow emphasizes state change rather than static geometry.
This supports our motivation for using dynamic 3D flow, which captures temporally varying scene dynamics more directly tied to action. For a fair comparison, all supervision baselines are implemented with the same auxiliary-head capacity and training protocol as FlowVLA.

\section{REAL-WORLD EXPERIMENTS}

\paragraph{Settings.}
We construct a real-world deployment setup with a UR5e platform equipped with third-person and wrist-view cameras. We evaluate four long-horizon manipulation tasks---\textit{Cup \& Plush}, \textit{Fruit}, \textit{Stack Cups}, and \textit{Cabinet}---and three more challenging tasks that require stronger 3D spatial understanding---\textit{Hang Cup}, \textit{Basketball}, and \textit{Ring Insertion}.
Zero-shot generalization is also evaluated under \textbf{viewpoint} and \textbf{height variations}. 
Further details on the setup are provided in Appendix~\ref{appendix:real_world_setup}.

\tabrealdeploy

\paragraph{Real-world Results.}
To validate real-world manipulation performance, we evaluate FlowVLA on four long-horizon base tasks and three 3D-aware tasks. In Tab.~\ref{table:realdeploy}, FlowVLA consistently improves baseline performance across the base tasks; notably, on \textit{Stack Cups}, FlowVLA fails only once out of 15 trials, demonstrating reliable multi-step manipulation.
The performance gain becomes more pronounced in the 3D-aware tasks that require precise spatial reasoning and action-relevant 3D dynamics. FlowVLA improves \textit{Basketball} by approximately 20\%p over the baseline and substantially boosts \textit{Hang Cup}  from 46.7\% to 86.7\%. These results indicate that scene-to-control 3D flow supervision effectively enhances control-relevant 3D scene understanding in real-world manipulation.
\figRealsetting

\tabrealvariation
\paragraph{Results under Test-Time Variations.}
We further evaluate zero-shot robustness under test-time variations. For \textit{Basketball}, we adjust the angle and height of the third-person camera; the initial ring height is modified for \textit{Ring Insertion}. As shown in Tab.~\ref{table:variation}, the baseline fails under viewpoint variation with 0\% success, whereas FlowVLA achieves 46.7\%. Under height variation in \textit{Ring Insertion}, FlowVLA improves the success rate from 46.7\% to 73.3\%, with a reduction in gripper collisions and inaccurate insertions. These results demonstrate that FlowVLA improves both in-distribution performance and zero-shot robustness against real-world spatial variations. A detailed failure case analysis is provided in Appendix~\ref{appendix:failure_case}.

\section{CONCLUSION}
We present \textbf{FlowVLA}, a reconstruction-guided 3D flow learning framework that injects action-relevant 3D dynamics into existing VLA models through training-only supervision.
By deriving temporally aligned 3D point trajectories from RGB observations into future 3D flow targets, FlowVLA couples scene-level point motion with control-relevant TCP displacement through dual-level flow supervision.
Experiments on LIBERO, LIBERO-Plus, and real-world tasks show consistent improvements in performance and robustness across different VLA backbones.
FlowVLA notably operates without auxiliary 3D sensors, architectural modifications, or inference-time computation, as all 3D flow modules are used only during training.
These results suggest that proposed 3D flow supervision serves as an effective approach to impart action-relevant 3D dynamics to 2D VLA models.

\paragraph{Limitations.}
FlowVLA requires an offline reconstruction step to generate 3D point trajectories for training. 
Although this does not affect inference-time efficiency, the quality of the auxiliary supervision depends on the reconstruction model. 
In addition, while the visualization in Appendix~\ref{appendix:selectedpoint} suggests that magnitude-based point selection tends to capture dynamically active regions, it does not explicitly guarantee task relevance. This suggests future extensions with semantic or contact-aware point selection.

\acknowledgments{This research was supported by the National Research Foundation of Korea (NRF), Electronics and Telecommunications Research Institute (ETRI), and Institute of Information \& Communications
Technology Planning \& Evaluation (IITP), funded by the Korean government [26CS1100, Development of Proprietary Physical AI-based Small-scale Computers and Integrated Soft Suits] and Korean government (MSIT) (RS-2026-25518808 and RS-2026-25617480).}

\bibliography{main}  

\clearpage
\appendix

\section{SIMULATION ENVIRONMENTS}\label{appendix_simulation_setting}
\subsection{Datasets}
\paragraph{LIBERO.} LIBERO~\cite{liu2023libero} is a standardized benchmark for robot manipulation. It consists of 4 task suites---LIBERO-Spatial, LIBERO-Object, LIBERO-Goal, and LIBERO-Long---where each suite contains 10 tasks and 50 expert demonstrations per task. These suites are designed to evaluate complementary aspects of manipulation capability, including spatial reasoning, generalization to unseen object instances, language-conditioned manipulation, and long-horizon planning.

\paragraph{LIBERO-Plus.} LIBERO-Plus~\cite{libero_plus} extends LIBERO to provide a more fine-grained evaluation of robustness under distribution shifts. It introduces 7 perturbation dimensions and contains 10,030 test episodes organized across 5 difficulty levels. These perturbations include changes in camera viewpoint, robot initialization, language instructions, lighting, background, layout, and sensor noise, thereby enabling a detailed analysis of zero-shot robustness beyond standard in-distribution evaluation.
\figliberoplus

\subsection{Ablation Study}
\tabreappendixtopk
\paragraph{Magnitude-based Top-$K$ Selection.}
We ablate the number of selected motion-salient 3D points under the same training setting by varying $K \in \{1024, 2048, 3072\}$.
As shown in Tab.~\ref{table:topk_ablation}, $K=2048$ achieves the best performance.
A smaller value, $K=1024$, may not provide sufficient coverage of scene-level motion, limiting the supervision of diverse motion-relevant dynamics.
Conversely, a larger value, $K=3072$, can include less informative regions such as static backgrounds or reconstruction noise, which weakens the quality of the flow supervision.
Nevertheless, FlowVLA maintains strong performance across all tested values of $K$, indicating that the method is not highly sensitive to this hyperparameter.
Thus, $K=2048$ offers the best trade-off between capturing meaningful 3D motion and filtering out irrelevant points.

\tabreappendixvggt
\paragraph{Effect of Reconstruction Model for 3D Flow Supervision.}
To examine the dependency of FlowVLA on the reconstruction model, we replace VDPM~\cite{vdpm} with VGGT~\cite{wang2025vggt} and another temporally consistent dynamics tracker, SpatialTrackerV2~\cite{xiao2025spatialtrackerv2}, for generating 3D flow supervision while keeping the policy architecture, training protocol, and evaluation setting unchanged. All 3D flow supervision settings improve over the baseline, indicating that FlowVLA is not specific to VDPM. However, VGGT performs lower than VDPM- and SpatialTrackerV2-based supervision. We attribute this gap to the fact that VDPM and SpatialTrackerV2 are explicitly designed for temporally consistent dynamic point trajectories, which provide cleaner motion targets for action-relevant 3D flow learning.

\tabreappendixgaussiannoise
\paragraph{Sensitivity to Noisy 3D Flow Supervision.}
To evaluate the robustness of FlowVLA to imperfect reconstruction targets, we conduct a noise sensitivity analysis on LIBERO-Long. After selecting $K$ motion-salient sparse points, we randomly corrupt a fixed percentage of selected point trajectories by adding zero-mean Gaussian noise to their 3D coordinates. As shown in Tab.~\ref{table:gaussian_noise}, performance decreases as the corruption ratio increases, from 87.8\% without noise to 84.7\% and 84.0\% under 5\% and 20\% corruption, respectively. Nevertheless, FlowVLA still outperforms the baseline of 80.4\% under both settings, indicating that the proposed supervision is not overly sensitive to moderate reconstruction noise. These results also confirm that the quality of reconstructed 3D trajectories remains an important factor for effective flow supervision.

\paragraph{Offline Reconstruction Cost.}
For Seer~\cite{seer} experiments on LIBERO-Long, VDPM~\cite{vdpm} preprocessing takes approximately 4 wall-clock hours to generate 3D flow supervision, compared to approximately 13 wall-clock hours for policy training excluding this preprocessing.
This offline cost is about 30\% of the policy training time, is incurred only once before training, and the generated supervision targets are cached for subsequent experiments.
Since the 3D flow supervision is training-only, it does not introduce any additional inference-time computation.

\section{REAL-WORLD ENVIRONMENTS}\label{appendix:real_world_setup}
\subsection{Real-World Setups}
As illustrated in Fig.~\ref{figure:real_deploy}, our real-world experimental setup consists of a single UR5e robotic arm equipped with two RealSense D435i cameras. The real-world evaluation comprises four long-horizon base tasks and three 3D-aware task scenarios. To assess the zero-shot generalization capability of VLA models, we additionally evaluate the models under viewpoint and height variations. For the Basketball task, we build the task environment using a Bambu Lab P1S 3D printer.
For collecting training demonstrations, we perform eye-to-hand camera calibration to obtain calibrated third-person observations and normalize all coordinates with respect to the robot base frame.

\figdeploysetting
\tabdeploydetail
\subsection{Datasets}
\figtaskwisefirst
\figtaskwisesecond
\paragraph{Base Tasks.}
We define four long-horizon manipulation tasks as base tasks.
\textbf{(a) Fruit} requires the robot to place a green apple into a pink bowl and an orange into a green bowl.
\textbf{(b) Cup \& Plush} requires the robot to place a white cup on top of an upside-down pink cup, and then put a blue penguin plush toy into the white cup.
\textbf{(c) Stack Cups} requires the robot to stack cups of different sizes in a specified order.
\textbf{(d) Cabinet} requires the robot to open the cabinet, swap the object inside with the object outside, and then close the cabinet.

\paragraph{3D-Aware Tasks.} We further design three 3D-aware tasks to assess whether the VLA model can capture action-relevant 3D dynamics and spatial relationships beyond conventional pick-and-place skills. \textbf{(a) Basketball} requires the robot to grasp a basketball and place it into a basketball hoop. \textbf{(b) Hang Cup} requires the robot to grasp a cup and accurately hang its handle on a cup holder. Unlike the base tasks, these scenarios require fine-grained 3D understanding beyond basic pick-and-place manipulation. \textbf{(c) Ring insertion} requires the robot to pick up two rings and precisely insert them onto a vertical pillar by aligning the ring holes with the pillar. We additionally evaluate zero-shot robustness under \textit{height} and \textit{viewpoint variations}, where the model is tested under shifted object heights and camera viewpoints without additional fine-tuning.

\subsection{Further Analysis}
\tabsupplecomponent
\paragraph{Component Analysis.}
We further conduct a component analysis under the \textit{viewpoint} variation setting. 
Removing coordinate alignment eliminates the robot-centric alignment of the 3D targets, while removing magnitude-based selection replaces motion-salient point selection with random sampling. 
We also ablate the point-level and TCP-level losses, which supervise scene-level motion and control-relevant motion, respectively. 
As shown in Tab.~\ref{tab:component_viewpoint}, all components contribute to robustness under viewpoint variation. In particular, removing magnitude-based selection or the TCP-level loss reduces the success counts from 7/15 to 0/15 and 2/15, respectively.

\paragraph{Real-World Deployment with $\pi_{0.5}$.}
\tabrealpi
We further evaluate FlowVLA with the strong baseline~\cite{pi05} on real-world 3D-aware tasks. 
As shown in Tab.~\ref{table:realpi}, FlowVLA improves the success rate on both Basketball and Hang Cup, from 10/15 to 13/15 and from 9/15 to 14/15, respectively. 
We additionally evaluate robustness to a dynamic visual environment by introducing a moving background during deployment in the Basketball task. 
Under this test-time dynamic-background variation, FlowVLA improves the success rate from 5/15 to 8/15, demonstrating improved robustness to dynamic scene changes beyond the static training environment.

\section{Qualitative Results}\label{appendix:visualization}

\subsection{3D Point Composition Analysis}
Fig.~\ref{figure:3dpoints} illustrates how our motion-based selection and TCP aggregation change the composition of 3D points in LIBERO-Spatial. Here, object points include robot components and scene objects, whereas non-object points correspond to the background. 
(\textbf{Panel}-\textcolor{red}{\ding{182}}) shows the 3D points labeled in the MuJoCo simulator, comprising 50k points. 
(\textbf{Panel}-\textcolor{red}{\ding{183}}) shows that the proportion of 3D object points, such as grippers and robot arms, becomes substantially higher than that of non-object points, such as the background, after applying our first-stage motion-based selection. 
We also demonstrate that the second selection stage in the TCP branch further filters out non-object 3D points during the attention operation, as shown in (\textbf{Panel}-\textcolor{red}{\ding{184}}). 
These results illustrate that motion-based selection suppresses largely static background regions, while TCP aggregation further emphasizes points associated with robot-object interactions.
\rebfig

\subsection{VDPM Visualization}
Fig.~\ref{figure:vdpm_visualize} and Fig.~\ref{figure:vdpm_visualize_libero} show qualitative visualizations of VDPM~\cite{vdpm} 3D point trajectories in LIBERO simulation and real-world scenarios. For visual clarity, we display all reconstructed points in this analysis, whereas FlowVLA uses only the top-$K$ dynamically informative points selected by the flow magnitude during training. The visualization confirms that VDPM can recover temporally consistent 3D point motion from RGB observation sequences. The resulting trajectories highlight motion-salient physical changes, including object displacement and interaction regions around the robot end-effector. These results provide qualitative evidence that reconstruction-derived 3D point trajectories can serve as a useful training-only supervision signal for FlowVLA.
\figvdpmvisualize
\figvdpmvisualizelibero

\paragraph{Selected Point Visualization.}\label{appendix:selectedpoint}
We provide a qualitative analysis of the proposed magnitude-based point selection by visualizing the selected $K$ points used for training-time flow supervision. 
As shown in Fig.~\ref{figure:magnitude_viz}, the selected points consistently appear around regions undergoing physical motion, such as the TCP, the falling object (the juice in the first frame), and the object interacting with the TCP (the basket in the sixth frame).
In contrast, points on static backgrounds and reconstruction-induced noisy regions are largely filtered out. 
These suggest that motion magnitude serves as a simple yet effective criterion for extracting motion-salient 3D points from reconstruction-derived trajectories, which often coincide with action-relevant regions.
Importantly, this selection does not require semantic masks, object annotations, or contact labels, making it a scalable mechanism for constructing 3D dynamics supervision from RGB-only demonstrations.

\figmagnitudefiltering

\paragraph{TCP Trajectory Visualization.}
We also visualize the predicted TCP trajectories to assess whether FlowVLA can convert scene-level point motion into control-relevant dynamics. Specifically, we compare the TCP trajectories predicted by the TCP-level aggregation module with the ground-truth TCP trajectories from robot execution. As shown in Fig.~\ref{figure:tcp_trajectory_viz}, the predicted trajectories closely follow the ground-truth trajectories, exhibiting similar motion directions. This qualitative agreement supports our scene-to-control distillation design: the model learns to aggregate dense 3D point displacements into an end-effector-level motion representation. Therefore, the auxiliary TCP-level objective provides a direct bridge between reconstruction-derived scene dynamics and the robot control trajectory.
\figtcptrajectory

\paragraph{TCP-Level Aggregation Weight Visualization.}\label{appendix:aggregationweight}
We further visualize the learned aggregation weights of the TCP-level branch to examine how FlowVLA transfers scene-level 3D motion cues into control-relevant dynamics. 
As shown in Fig.~\ref{figure:tcpaggregate_viz}, magnitude-based selection identifies $K$ motion-salient 3D points from reconstruction-derived trajectories, and we visualize the top-512 points with the largest TCP-level aggregation weights. 
The TCP-level aggregation module further refines these scene-level motion cues by assigning larger weights to points around the robot TCP and object--robot interaction region. 
This indicates that the TCP-level branch learns to distill motion-salient 3D point displacements into end-effector-relevant flow, rather than treating all selected points uniformly. 
This visualization further supports our scene-to-control distillation design by showing how the TCP-level branch extracts control-relevant motion cues from reconstruction-derived scene dynamics.
\figaggregatedweight

\subsection{Failure Case Visualization} \label{appendix:failure_case} 
Fig.~\ref{figure:failurecase_basic_part2}, Fig.~\ref{figure:failurecase_basketball}, and Fig.~\ref{figure:failurecase_ring} visualize representative failure cases in the real-world experiments. With limited demonstration data, the baseline frequently stops execution prematurely or places objects at incorrect locations under variations in the initial object configuration. These failures become more severe in 3D-aware tasks and \textbf{viewpoint variation}, often leading to collisions or emergency stops. In the \textbf{height variation} setting, the baseline also struggles to recognize shifted grasp positions, resulting in missed grasps or collisions with the table surface. FlowVLA substantially mitigates these failure modes. It shows no premature stopping behavior and performs precise manipulation in most 3D-aware scenarios, except for extreme cases where the target object lies near the camera boundary. FlowVLA also achieves more accurate grasping under \textbf{viewpoint} and \textbf{height variations}, significantly reducing collisions and insertion failures. These qualitative results suggest that training-only 3D flow supervision improves the 3D spatial awareness of VLA policies, even with limited demonstration.
\figRealcasesecond
\figRealcasebasket
\figRealcasering

\end{document}